\documentclass[lettersize,journal]{IEEEtran}
\usepackage{amsmath,amsfonts}
\usepackage{algorithmic}
\usepackage{algorithm}
\usepackage{array}
\usepackage[caption=false,font=normalsize,labelfont=sf,textfont=sf]{subfig}
\usepackage{textcomp}
\usepackage{stfloats}
\usepackage{url}
\usepackage{verbatim}
\usepackage{graphicx}
\usepackage{cite}
\usepackage{booktabs}
\usepackage{makecell}
\usepackage{afterpage}
\usepackage{multirow}
\usepackage{breakcites}
\usepackage[caption=false]{subfig}
\usepackage{siunitx}
\usepackage{xcolor}
\usepackage{chngcntr} 
\usepackage{subfiles} 
\graphicspath{{./figs/}}     
\newif\ifsupplement
\supplementfalse 

\begin{document}

\title{MLLM-CTBench: A Benchmark for Continual Instruction Tuning with Reasoning Process Diagnosis}

\author{%
Haiyun Guo, Zhiyan Hou, Yandu Sun, Jinghan He, Yu Chen, Yuzhe Zhou, Yuheng Jia, Jinqiao Wang, and Tat-Seng Chua%
\thanks{Haiyun Guo, Zhiyan Hou, Jinghan He and Jinqiao Wang are with the Institute of Automation, Chinese Academy of Sciences, Beijing 100190, China
(e-mail: haiyun.guo@nlpr.ia.ac.cn; houzhiyan23@mails.ucas.ac.cn; hejinghan2022@ia.ac.cn;  jqwang@nlpr.ia.ac.cn).}%
\thanks{Yandu Sun, Yuzhe Zhou , Yu Chen and Yuheng Jia are with Southeast University, Nanjing 211189, China (email: 220252288@seu.edu.cn; 2132230472@seu.edu.cn ; yu.chen.8525@gmail.com ;  yhjia@seu.edu.cn).}%
\thanks{Tat-Seng Chua is with the School of Computing, National University of Singapore, Singapore 117417
(e-mail: dcscts@nus.edu.sg).}%

}

\maketitle




\maketitle

\begin{abstract}
Continual instruction tuning(CIT) during the post-training phase is crucial for adapting multimodal large language models (MLLMs) to evolving real-world demands. However, the progress is hampered by the lack of benchmarks with rigorous, protocol-consistent evaluation. To bridge this gap, we introduce MLLM-CTBench, a comprehensive benchmark for CIT of MLLMs, covering seven challenging tasks across six diverse domains. MLLM-CTBench makes three key contributions. 
First, we establish a multidimensional evaluation framework that jointly assesses final-answer accuracy and process-level reasoning quality, where Chain-of-Thought (CoT) traces serve as an observable signal to diagnose catastrophic forgetting beyond answer-only evaluation.
Second, we conduct a large-scale evaluation of continual learning methods by systematically assessing eight representative algorithms from four major families under a unified protocol across task orders, providing actionable insights for algorithm design. Third,
we expand the scope from Supervised Fine-Tuning (SFT) to Reinforcement Fine-Tuning (RFT) in CIT.
By investigating GRPO, an on-policy RL algorithm that stabilizes updates through explicit KL-divergence control to a prior policy, we aim to analyze how this mechanism affects cross-task knowledge retention.
Our experiments yield several findings: 
(1) Process-level reasoning quality is often more resilient to catastrophic forgetting than final-answer accuracy, and forgetting is primarily driven by degradation in domain knowledge. (2) Model capability is critical factor influencing continual learning outcomes, with stronger baseline models exhibiting greater resistance to catastrophic forgetting. (3) On-policy RFT (GRPO), with its inherent KL control, achieves more stable cross-task retention than SFT. While removing KL control can amplify forgetting despite potential gains on new ones. 
We release the code and dataset to facilitate reproducibility.\footnotemark
\end{abstract}
\footnotetext{Code: \url{https://github.com/yueluoshuangtian/MLLM-CTBench}. 
Dataset: \url{https://huggingface.co/datasets/yueluoshuangtian/MLLM-CITBench}.}

\begin{figure*}[t]
  \centering
  \includegraphics[width=0.9\textwidth]{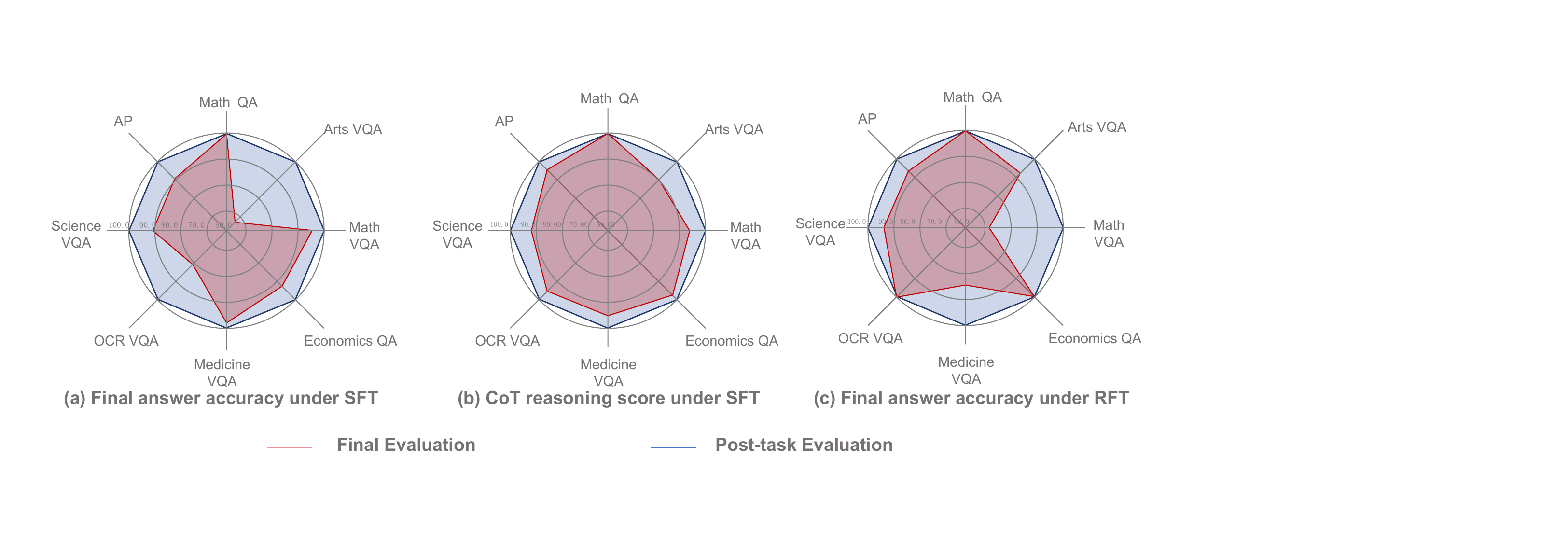}
    \caption{\textbf{CIT evaluation of MLLMs under SFT/RFT with reasoning process diagnosis.}
We denote by $P_{j,j}$ the post-task performance on task $j$ measured immediately after training on task $j$, and by $P_{N,j}$ the final performance after completing training on all tasks. We normalize post-task performance to 100 (blue) and report normalized final performance as $P_{N,j}/P_{j,j}\times100$ (red). Values below 100 indicate forgetting. Compared to panel (a), the smaller deviations from the post-task reference in (b) indicate that CoT reasoning quality degrades more slowly than final-answer accuracy.
Comparing (a) and (c), RFT(GRPO) exhibits overall higher normalized final performance across tasks, demonstrating more effective knowledge retention in MLLMs than SFT in CIT.
}

  \label{fig:abstract}
\end{figure*}
\section{Introduction}
\IEEEPARstart{M}{ultimodal}  Large Language Models (MLLMs) have emerged as foundational architectures for cross-modal understanding and generation, demonstrating impressive capabilities across a variety of tasks. Instruction tuning has further enhanced these models by aligning them with human intent and improving task-specific performance through supervised adaptation~\cite{yu2024recentadvancesmultimodalcontinual}. However, real-world deployment demands continuous adaptation to evolving instructions and domain requirements—a paradigm known as \emph{continual instruction tuning}~\cite{he2023continual,CoIN}, where the model incrementally learns from new tasks while retaining prior capabilities. 

While significant progress has been made in continual instruction tuning for Large Language Models (LLMs)~\cite{freeze}, the multimodal counterpart still lags behind. A major obstacle is the lack of a rigorous and comprehensive evaluation benchmark. Existing benchmarks (e.g., EMT~\cite{EMT}, CITB~\cite{CITB}, CoIN~\cite{CoIN})  for continual instruction tuning of MLLMs exhibit critical limitations. 
(1)  \textbf{Insufficient evaluation granularity}: 
Most benchmarks primarily emphasize final answer correctness, offering limited visibility into the reasoning process and hindering causal diagnosis of catastrophic forgetting~\cite{Luo2023AnES}. 
Although CoIN~\cite{CoIN} implicitly estimates reasoning knowledge forgetting, the interpretability of the evaluation metric remains inadequate. 
(2)  \textbf{Limited Investigation of Continual Learning algorithms and Training Paradigms}: 
Existing works mainly quantify catastrophic forgetting under under sequential fine-tuning, while providing only limited coverage of continual learning algorithms, thereby constraining the practical utility and actionable insights of such benchmarks. Moreover, despite the increasing adoption of reinforcement fine-tuning (RFT) to improve MLLMs' reasoning and generalization, its stability and effectiveness in continual instruction tuning settings remain insufficiently characterized. 
(3) \textbf{Weak alignment with modern MLLM post-training objectives}:
Prevailing multimodal continual instruction benchmarks are largely built upon conventional short-answer datasets (e.g., classification-centric datasets in EMT~\cite{EMT}, and VQAv2/TextVQA/GQA/OCR-VQA with limited reasoning tasks in CoIN~\cite{CoIN}), which only weakly reflect today’s post-training emphasis on multi-step, Chain-of-Thought--style visual reasoning.
As a result, they offer limited headroom on some tasks and, more importantly, limited diagnostic power for characterizing capability degradation in modern MLLMs under continual tuning.
This motivates benchmarks that include reasoning-intensive instructions with process-level evaluation.

To catalyze research progress in continual instruction tuning for MLLMs, we introduce MLLM-CTBench, a benchmark tailored to the limitations outlined above. Our design is guided by three goals: (i) Process-level diagnosis beyond answer-only scoring, (ii) Reasoning-inclusive, modern post-training–aligned task design, and (iii) Protocol-consistent benchmarking for fair comparison across methods, paradigms and task orders. Concretely, we make three contributions.
\textbf{(1) Multi-dimensional evaluation with reasoning process diagnosis.} Moving beyond answer-only evaluation, MLLM-CTBench jointly measures final-answer correctness and the quality of the intermediate reasoning process, enabling fine-grained, process-level diagnosis of catastrophic forgetting under CIT. 
Specifically, we train a dedicated MLLM-based evaluator that produces interpretable, fine-grained scores along complementary dimensions, including visual grounding fidelity (for VQA-style tasks), logical coherence, and domain-knowledge retention, which supports targeted analyses beyond aggregate accuracy~\cite{tan2024judgebench,zheng2023judging}. To mitigate subjectivity and model-specific bias, we calibrate the evaluator using consistency-filtered GPT-4 preference supervision together with GRPO-based refinement, and validate its agreement with human judgments across models and training paradigms (SFT/RFT) (Sec.~\ref{MLLM evaluator}).
 \textbf{(2) Reasoning-inclusive benchmark with systematic method coverage.}
MLLM-CTBench curates approximately 70K high-quality instances from 16 public datasets, spanning seven tasks across six domains (Math, OCR, Science, Medicine, Arts, and Economics). The benchmark is constructed to better match modern post-training objectives by emphasizing reasoning-intensive instructions, OCR robustness, and domain knowledge demands~\cite{iconqa,chen2022geoqageometricquestionanswering,xia2024chartx,yue2023mmmu,estvqa,ScienceQA,vqarad,ImageCLEF-VQA-Med2021,he2020pathvqa,zhang2023pmcvqa,garcia2020AQUA,trace}.
Under a unified protocol, we evaluate eight representative continual learning algorithms from four major families: regularization-based~\cite{ewc,freeze,lwf,mas}, replay-based~\cite{replay,der,wang2024relational}, architecture-based~\cite{l2p,cui2025cmoa}, and model-fusion--based approaches~\cite{MagMaX}. 
\textbf{(3) Comparative Analysis of SFT and RFT under continual instruction tuning}.
While prior analyses in CIT have primarily focused on SFT, we extend the scope to include on-policy RFT algorithms (e.g., GRPO~\cite{grpo,zhang2025reinforcement}), which adopts an explicit KL-divergence constraint to mitigate drastic policy drift during sequential adaptation. We empirically investigate how this  mechanism of on-policy RL influences cross-task knowledge dynamics.


Leveraging MLLM-CTBench, we uncover several findings under the above controlled protocols.  
\textbf{(i) Stronger base MLLMs are generally more resistant to catastrophic forgetting}. Our large-scale evaluation confirms that the capacity of the base MLLM is a critical factor influencing outcomes, with stronger models exhibiting greater resistance to catastrophic forgetting.
\textbf{(ii) Process-level signals are relatively resilient and diagnostic}. Compared with final-answer accuracy, CoT-quality metrics often exhibit smaller post-task-to-final degradation (Fig.\ref{fig:abstract}(a,b)), while answer-level forgetting aligns most strongly with knowledge degradation among the evaluated CoT dimensions, supporting a hierarchical and interpretable pattern of forgetting~\cite{CoIN,zheng-etal-2024-learn,zheng2025spuriousforgettingcontinuallearning}.
\textbf{(iii) On-policy RFT reduces forgetting more effectively than SFT in CIT}.  Our results confirm that on-policy RFT better maintains cross-task retention (Fig.\ref{fig:abstract}(c))  than SFT (Fig.\ref{fig:abstract}(a)). And the KL constraint in RFT is critical for mitigating forgetting. This evidence that constrained policy updates enhance retention is in line with concurrent analyses of on-policy RL~\cite{shenfeld2025rl}.
\textbf{(iv) Algorithm effectiveness is regime-dependent}. Replay-based methods tend to benefit weaker models more, regularization-based approaches are more competitive with higher-capacity models, and model fusion often provides a favorable trade-off between retention and efficiency under resource constraints.

\begin{figure*}[t]
  \centering
  \includegraphics[width=0.9\linewidth]{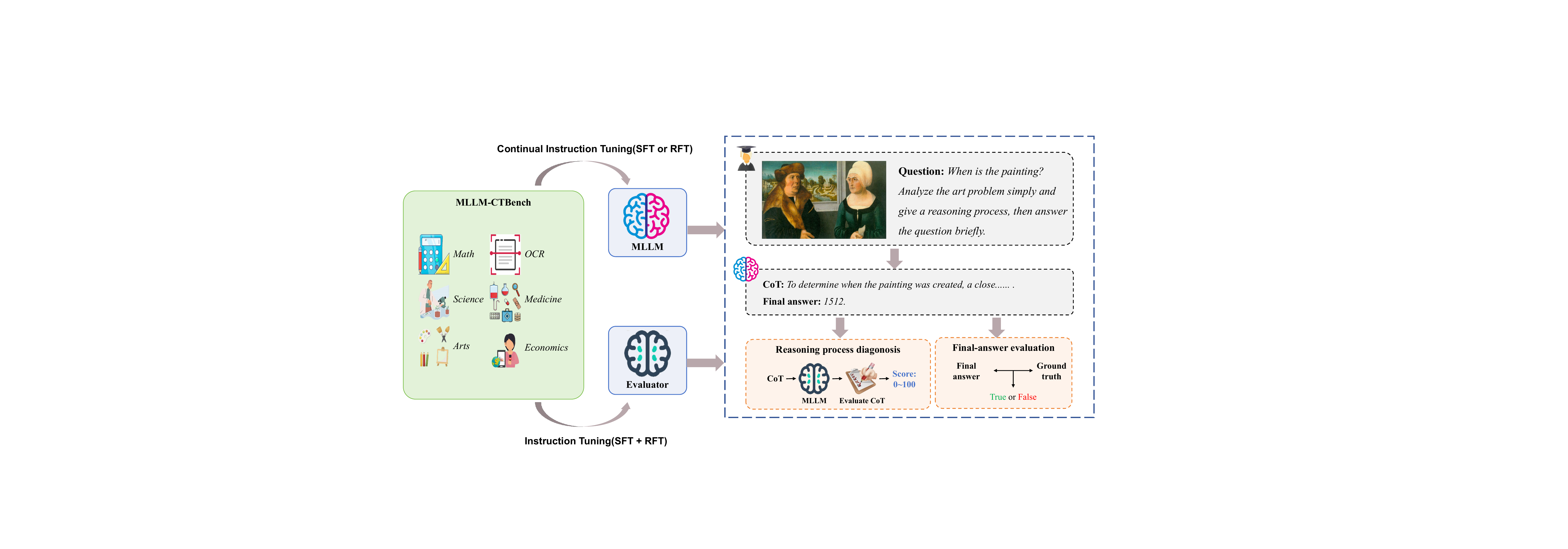}
\caption{Overview of \textbf{MLLM-CTBench}. The MLLMs firstly undergo continual instruction tuning on a sequence of tasks curated from six diverse domains. Then the performance is measured under multidimensional evaluation metrics combining both final answer evaluation with reasoning process diagnosis (including visual grounding, logical coherence, and knowledge retention) enabled by a dedicated MLLM evaluator.}
  \label{fig:introduction}
\end{figure*}

\section{Related Work}

\subsection{Method Design for Continual Instruction Tuning}
Continual instruction tuning (CIT) incrementally adapts (M)LLMs to new instruction distributions while retaining prior capabilities~\cite{he2023continual,CoIN}. Existing approaches largely follow the continual learning (CL) taxonomy~\cite{wu2024continual}. \textbf{(1) Regularization-based} methods constrain updates to preserve past knowledge, such as EWC~\cite{Kirkpatrick_2017}, OGD~\cite{farajtabar2020orthogonal}, LwF~\cite{li2017learning}, and MAS~\cite{mas}. \textbf{(2) Replay-based} methods revisit stored or synthesized samples to reduce forgetting~\cite{rolnick2019experience,replay,der,wang2024relational}, at cost of memory or compute. \textbf{(3) Architecture-based} methods add task-conditioned components (e.g., progressive expansion or module routing)~\cite{razdaibiedina2023progressive,l2p,cui2025cmoa}, alleviating interference but potentially increasing inference overhead. \textbf{(4) Model-fusion-based} strategies merge task-specific checkpoints post hoc to obtain a single model with limited extra training~\cite{MagMaX}. Despite progress, multimodal CIT remains harder due to cross-modal alignment drift and diverse visual reasoning demands, calling for systematic evaluation tailored to modern MLLM post-training.

\subsection{Benchmarks for Continual Instruction Tuning}
Benchmarks are essential for diagnosing catastrophic forgetting. For LLMs, CITB~\cite{CITB} provides standardized continual instruction protocols but is text-centric and often outcome-focused. For MLLMs, EMT~\cite{EMT} highlights multimodality degradation under fine-tuning, while CoIN~\cite{CoIN} offers a representative continual instruction benchmark. However, existing MLLM benchmarks typically (i) emphasize final-answer correctness, (ii) rely heavily on short-answer VQA-style tasks, and (iii) under-characterize training paradigms beyond SFT. In particular, reinforcement fine-tuning (RFT) is increasingly used in post-training ~\cite{ouyang2022training,Chung2022ScalingIL}, yet their stability in continual settings remains insufficiently benchmarked.

\subsection{CoT Reasoning Evaluation and LLM/MLLM-as-a-Judge}
LLM-based judges provide scalable alternatives to human evaluation via rubric-guided scoring, pairwise comparison, and preference calibration~\cite{zhu2023judgelm,li2023generative,bai2023touchstone,kim2023prometheus}. These ideas extend to vision-language models: MLLM-as-a-judge supports scoring and ranking in multimodal evaluation~\cite{chen2024mllmasajudgeassessingmultimodalllmasajudge}, where grounding fidelity and hallucination become central. In CIT, incorporating process-level signals (e.g., CoT traces) can improve interpretability beyond answer-only metrics by revealing \emph{how} performance degrades under sequential adaptation. This motivates CoT-aware and calibrated MLLM evaluators that provide fine-grained, dimension-wise diagnostics (e.g., grounding fidelity, logical coherence, and domain-knowledge consistency) for evaluation.



\section{MLLM-CTBench}

We design MLLM-CTBench to evaluate continual instruction tuning of MLLMs under modern, reasoning-oriented post-training objectives. Our construction follows two principles:
\textbf{(i) Multi-dimensional evaluation}: in addition to final-answer correctness, we assess process-level CoT quality using a dedicated evaluator, enabling more interpretable analyses of forgetting and capability drift. 
\textbf{(ii) Task curation for continual scenarios}: we curate a sequential task suite spanning diverse domains with non-saturated difficulty and clear domain shifts, so that both continual adaptation and cross-task retention can be meaningfully measured.

\subsection{Benchmark Construction}

\subsubsection{Domains and tasks}
MLLM-CTBench targets continual instruction tuning under modern, reasoning-intensive post-training objectives. We focus on six performance-limited domains---\textbf{Arts}, \textbf{Medicine}, \textbf{Economics}, \textbf{Science}, \textbf{Math}, and \textbf{OCR}---where current MLLMs still exhibit substantial headroom. For instance, strong models obtain only 51.9\% on MMMU-Pro~\cite{yue2024mmmuprorobustmultidisciplinemultimodal} and up to 61.5\% on OCRBench v2~\cite{fu2024ocrbenchv2improvedbenchmark}. 
We organize the benchmark into \textbf{7 tasks} spanning these domains: \textit{Math QA}, \textit{Economics QA}, \textit{Science VQA}, \textit{Math VQA}, \textit{Medicine VQA}, \textit{OCR VQA}, and \textit{Arts VQA}. Examples of each task are given in Fig.\ref{fig:supp_examples} in Supplementary. This task suite jointly stresses symbolic reasoning, domain knowledge retention, and visually grounded inference, making it suitable for diagnosing catastrophic forgetting beyond short-answer VQA.

\textbf{Curation criteria.} To better reflect continual learning challenges, we curate instances using four principles:  \textbf{(a) reasoning-intensive instructions} (multi-step inference rather than short look-up),  \textbf{(b) strong visual dependency }(solutions require image evidence for VQA tasks),  \textbf{(c) non-saturated difficulty}, and \textbf{(d) clear domain shifts} to stress cross-task retention.
We cap the number of training instances per task to a similar scale and report task-wise results to avoid dominance by large tasks.



\subsubsection{Data Integration}

We construct MLLM-CTBench from public datasets across six domains:
\textbf{(1) Arts} from AQUA~\cite{garcia2020AQUA};
\textbf{(2) Science} from ScienceQA~\cite{ScienceQA} and AI2D~\cite{ai2d};
\textbf{(3) Medicine} from VQA-RAD~\cite{lau2018dataset}, VQA-Med~\cite{ImageCLEF-VQA-Med2021}, PMC-VQA~\cite{zhang2023pmcvqa}, and PathVQA~\cite{he2020pathvqa};
\textbf{(4) Economics} from TRACE~\cite{wang2023tracecomprehensivebenchmarkcontinual};
\textbf{(5) Math} from IconQA~\cite{lu2021iconqa}, GeoQA~\cite{chen2022geoqageometricquestionanswering}, CHARTX~\cite{xia2024chartx}, MMMU~\cite{yue2023mmmu}, and TRACE;
\textbf{(6) OCR} from ChartOCR~\cite{chatocr}, CROHME~\cite{guan2024posformerrecognizingcomplexhandwritten}, and ESTVQA~\cite{wang2020generalvalueevidencebilingual}.
Dataset statistics are summarized in Table~\ref{Task_Composition}.


\begin{table}[ht]
    \centering
    \caption{Statistics of the MLLM-CTBench datasets.} 
    \resizebox{\columnwidth}{!}{
        \begin{tabular}{cccc}
            \toprule
            \textbf{Task} &
            \textbf{Data Source} &
            \makecell[c]{\textbf{Train} \\ \textbf{(Text / Image)}} &
            \makecell[c]{\textbf{Test} \\ \textbf{(Text / Image)}} \\
            \midrule
            \rule{0pt}{1em}\textbf{Math QA} & \makecell[c]{TRACE} & 10K/0 & 0.5K/0 \\  \hline
            \rule{0pt}{1em}\textbf{Economics QA} & \makecell[c]{TRACE} & 5K/0 & 0.5K/0 \\ \hline
            \rule{0pt}{1em}\textbf{Science VQA} & \makecell[c]{AI2D, ScienceQA} & 9K/4K & 1K/0.5K \\ \hline
            \rule{0pt}{1em}\textbf{Math VQA} & \makecell[c]{IconQA, GeoQA, CHARTX, \\ MMMU} & 8.3K/8.3K & 0.9K/0.9K \\ \hline
            \rule{0pt}{1em}\textbf{Medicine VQA} & \makecell[c]{VQA-RAD, VQA-Med-2021, \\ PMC-VQA, PathVQA} & 9K/6.9K & 1K/1K \\ \hline
            \rule{0pt}{1em}\textbf{OCR VQA} & \makecell[c]{ChartOCR, CROHME, \\ ESTVQA} & 12K/12.1K & 1.4K/1.4K \\ \hline
            \rule{0pt}{1em}\textbf{Arts VQA} & \makecell[c]{AQUA} & 9K/7K & 1K/0.9K \\ 
            \bottomrule
        \end{tabular}
    }
    \label{Task_Composition}
\end{table}

\subsubsection{CoT Annotation}
We provide CoT-style supervision for training instances to better match reasoning-intensive post-training~\cite{zhang2023multimodal}. Since tasks vary in answer formats (multiple choice, open-ended, yes/no), we design \emph{task-} and \emph{format-specific} instruction templates and show the details in Tab.\ref{tab:supp_prompts_metrics} in Supplementary. 
Each input includes the problem statement, answer format constraints, and task-specific instructions.
We query GPT-4~\cite{OpenAI2023GPT4TR} with structured prompting~\cite{Liu2023PromptLT} to elicit explicit step-by-step rationales, which improves interpretability and supports process-level evaluation. Importantly, GPT-4 is only used for offline annotation and training, all benchmark evaluation is performed on model-generated outputs.

\subsection{Continual Instruction Tuning}

\textbf{Task orders.}
We conduct sequential finetuning under two task permutations:
\textbf{Order-A} (Math QA $\rightarrow$ Arts VQA $\rightarrow$ Math VQA $\rightarrow$ Economics QA $\rightarrow$ Medicine VQA $\rightarrow$ OCR VQA $\rightarrow$ Science VQA) and its reverse \textbf{Order-B}. Task sequencing can change the degree of interference because tasks differ in modality demands (e.g., OCR vs. VQA), in the balance between multi-step reasoning and domain knowledge, and in output formats. Such differences make transfer asymmetric: learning a task that strongly updates certain capabilities may disproportionately affect previously learned tasks that rely on overlapping representations. In this paper, we use Order-A and its reverse (Order-B) as a minimal but controlled probe of order sensitivity.

\textbf{Supervised Fine-tuning (SFT).}
Given tasks $\{\mathcal{T}_1,\dots,\mathcal{T}_N\}$ with datasets $\{\mathcal{D}_1,\dots,\mathcal{D}_N\}$, SFT optimizes $f_\theta$ on each task by minimizing
\begin{equation}
\mathcal{L}_{\mathcal{T}_i}
= \frac{1}{|\mathcal{D}_i|}\sum_{(x^{\text{img}},x^{\text{ins}},y)\in \mathcal{D}_i}
\ell\bigl(f_\theta(x^{\text{img}},x^{\text{ins}}),y\bigr),
\label{eq:sft}
\end{equation}
where $\ell$ is cross-entropy. We evaluate both full-parameter tuning and LoRA~\cite{hu2021loralowrankadaptationlarge} to cover different adaptation regimes.

\textbf{Reinforcement Fine-tuning (RFT) with GRPO.}
We further study GRPO, an on-policy RL algorithm. Let the \emph{state} be the conditioning input $s=(x^{\text{img}},x^{\text{ins}})$ and the \emph{action} be the generated answer sequence $a=y_{1:T}\sim \pi_\theta(\cdot|s)$.
GRPO optimizes

\begin{equation}
\mathcal{L}_{\text{GRPO}}
= \mathbb{E}_{(s,a)\sim \pi_{\theta_{\text{old}}}}
\Bigl[
r_\theta(s,a)\,A(s,a) - \beta\,\mathrm{KL}\!\bigl(\pi_\theta\,\|\,\pi_{\theta_{\text{ref}})}
\Bigr],
\label{eq:grpo}
\end{equation}
where $r_\theta(s,a)=\frac{\pi_\theta(a|s)}{\pi_{\theta_{\text{old}}}(a|s)}$ is the policy ratio and $\beta$ controls KL constraint.
We compute the advantage $A(s,a)$ from group-sampled candidates as a normalized reward signal, following GRPO practice~\cite{shao2024deepseekmath}.
The KL constraint explicitly controls policy drift, which is critical for stability under sequential adaptation.

\textbf{Joint SFT--RFT.}
Following the common practice in MLLM post-training, for each new task, we first run one epoch of SFT initialization and then apply GRPO-based RFT, serving as a practical hybrid that balances fast adaptation and stability.

\subsection{Multidimensional Evaluation Metrics}
\subsubsection{Final-Answer Evaluation} \label{macro-level}

We extract the final answer from outputs (composed of reasoning traces and final answer) and compare it to the ground truth using task-specific metrics. 
We utilize the final answer performance to compute two standard metrics in order to evaluate CIT of MLLM. 
Let $P_{i,j}$ denote the \emph{final-answer score} on task $j$ after training on task $i$, and $N$ be the total number of tasks.

\textbf{Average Performance (AP)} measures overall performance after all tasks are sequentially trained:  
$AP = \frac{1}{N} \sum_{j=1}^{N} P_{N,j}$. A higher AP indicates better task-wide performance.

\textbf{Backward Transfer (BWT)} quantifies the effect of new-task learning on prior tasks:  
$BWT = \frac{1}{N-1} \sum_{j=1}^{N-1} (P_{N,j} - P_{j,j})$. Negative BWT reflects forgetting, while positive values indicate beneficial transfer.

\subsubsection{Reasoning Process Diagnosis}
\label{micro-level}
To diagnose \emph{process-level} degradation beyond final-answer metrics, we evaluate the quality of model-generated CoT reasoning traces. 
We adopt two complementary evaluators: (i) a general-purpose open-source judge for transparent, prompt-based scoring, and (ii) a dedicated trained MLLM evaluator for improved cross-model and cross-paradigm consistency.

\textbf{General-Purpose Evaluator.}
Following judge-style evaluation protocols~\cite{ho2022large}, we use Qwen-VL-32B as an open-source MLLM judge~\cite{chen2024mllmasajudgeassessingmultimodalllmasajudge}.
Given the input $(x^{\text{img}}, x^{\text{ins}})$, the model's response (including its CoT trace and final answer), and the ground-truth answer(including reasoning trace generated by GPT-4 and answer annotation), the judge assigns three dimension-wise scores in $[0,100]$ for the reasoning process:
\textbf{(i) Logical Coherence}, 
\textbf{(ii) Visual Grounding Fidelity} (only for VQA-style tasks), and 
\textbf{(iii) Domain Knowledge Retention}.
To improve interpretability and reduce subjectivity, each dimension is assessed using a \emph{four-level rubric}
(\textit{Irrelevant}, \textit{Partially Correct}, \textit{Mostly Correct}, \textit{Fully Correct}),
which is mapped to a corresponding numeric score range.\footnote{We follow a fixed rubric-to-score mapping; see Supplementary for the exact rubric prompts and examples.}
In addition, we include a small set of \emph{anchor examples} in the system prompt, each containing the question, a candidate response, the reference answer, and per-dimension ratings, to stabilize scoring across tasks and answer formats.
Finally, we aggregate the applicable dimensions by averaging to obtain an overall CoT-quality score for each instance.\footnote{For non-VQA tasks, the grounding dimension is omitted and we average the remaining dimensions.}
The full prompts and illustrative scoring examples are provided in Supplementary (Fig.~\ref{fig:supp_rubric} and Fig.~\ref{fig:supp_eval_example}).

\textbf{Dedicated MLLM Evaluator.}
\label{MLLM evaluator}
To further improve scoring consistency across evaluated models and training paradigms, we train a dedicated evaluator based on Qwen2.5-VL-7B using a two-stage pipeline.
\textbf{Stage 1: SFT with consistency-filtered preference supervision.}
We collect multiple responses to the same query from different training stages and task orders, and construct \emph{pairwise comparison} samples.
GPT-4 provides pairwise preferences and brief rationales.
To increase supervision reliability, we apply \emph{consistency filtering}: we retain only responses whose preferences remain consistent across at least two independent pairings, and use the associated rationales as high-confidence supervision.
To mitigate position bias, we randomize the order of the two responses within each pair~\cite{dvivedi2024comparativeanalysislargelanguage}.
The resulting preference dataset contains \textbf{2100} pairs and \textbf{15} evaluated model checkpoints.
\textbf{Stage 2: GRPO-style RFT for preference optimization.}
We further refine the evaluator using GRPO-style preference optimization~\cite{zhang2024direct}, treating GPT-4 preference signals as rewards.
This second stage improves calibration and robustness when scoring outputs from different MLLMs and from different post-training paradigms (SFT/RFT), especially when output styles and verbosity vary across models.

\textbf{Human Alignment Checks.}
We assess evaluator reliability against human annotations on a stratified subset covering all domains/tasks and multiple evaluated models under both SFT and RFT.
We report Spearman's $\rho$, Pearson's $r$, and Kendall's $\tau$ to quantify score agreement (see Tab.~\ref{tab:eval_IAA} and Tab.~\ref{tab:eval_evaluator}).
For completeness, metric definitions are provided in Supplementary (Sec.~\ref{sec:supp_corr_metrics}).
The human-labeled subset contains \textbf{630} instances (with \textbf{90} per domain/task on average).

\section{Experiments}

\subsection{Experimental Setup}
\label{sec:exp_setup}

\noindent\textbf{Research Questions.}
Our experiments are designed to answer the following questions under a unified CIT protocol.
    \textbf{RQ1:} \emph{ 
    What additional insights does process-level diagnosis provide beyond answer-only metrics for evaluating catastrophic forgetting? }
    \textbf{RQ2:} \emph{ 
    How do representative continual learning method families (regularization, replay, architecture expansion, and model fusion) compare in terms of retention on MLLM-CTBench?}
    \textbf{RQ3:} \emph{   
    How does RFT (GRPO) compare with SFT under CIT, and what role does KL control play in cross-task retention?}
   \textbf{RQ4:} \emph{
    How sensitive are continual learning outcomes to base-model capability in our setting?}
    
We answer RQ1 by validating the CoT evaluator and analyzing final answer-only vs. CoT-based process-level forgetting; RQ2 by the main benchmark results across method families; RQ3 by controlled SFT/RFT comparisons and KL ablations; and RQ4 by experiments varying base models.

\begin{table}[t]
\centering
\caption{Inter-annotator agreement (IAA) among three human raters for different models and training paradigms. Values are correlations $\times 100$; higher indicates stronger consistency.}
\label{tab:eval_IAA}
\setlength{\tabcolsep}{2.6mm}
\renewcommand{\arraystretch}{0.95}
\resizebox{\columnwidth}{!}{%
\begin{tabular}{l l S S S}
\toprule
\textbf{Model} & \textbf{Training} 
& {\textbf{Spearman $\rho$}} & {\textbf{Pearson $r$}} & {\textbf{Kendall $\tau$}} \\
\midrule
Qwen2.5-VL-3B & SFT & 85.16 & 90.16 & 76.97 \\
Qwen2.5-VL-3B & RFT  & 87.86 & 93.69 & 74.29 \\
LLaVA-1.5-7B  & SFT & 86.48 & 92.56 & 79.86 \\
\bottomrule
\end{tabular}%
}
\end{table}

\textbf{Metrics.}
Throughout the paper, we report task performance in percentage points for macro-level final answers. 
For continual learning, we summarize per-task performance with:
(i) \textbf{Acc}: the final task performance after completing training on all tasks,
(ii) \textbf{Forget}: the change relative to the performance measured immediately after learning that task.
By definition, \textbf{Forget} is reported in percentage points: negative values indicate forgetting, while positive values indicate backward transfer.
We use ``--'' for the last task in a sequence.
We also report \textbf{AP} and \textbf{BWT} (refer to Sec.~\ref{macro-level}).

When measuring micro-level reasoning quality, we use evaluator-assigned \textbf{CoT-Score} in $[0,100]$ (refer to Sec.~\ref{micro-level}), and compute its Forget in the same way as above.
To avoid ambiguity, we denote evaluator results by \textbf{Score/Forget} and reserve \textbf{Acc/Forget} for final-answer metrics.

\textbf{Training protocol}.
We evaluate three open-source MLLMs: LLaVA-1.5-7B and Qwen2.5-VL-3B.
Unless otherwise specified, we fix the input preprocessing and maximum sequence length (4096 tokens) across all methods and orders.
For baseline comparison settings, we include:
(i) \textbf{Zero-shot}: evaluation without task-specific finetuning,
(ii) \textbf{Direct FT}: finetuning separately for each task,
(iii) \textbf{Multi-task}: joint training on the union of all tasks.

\textbf{Unified optimization settings}
To improve fairness and comparability, we keep training components consistent within each model family.
Concretely, for each backbone we fix the optimizer type and scheduling strategy across all continual methods; differences are limited to method-specific components (e.g., replay buffer, regularization penalty, prompt modules).
We report model-specific learning rate, batch size, and epochs below. 
\begin{itemize}

    \item \emph{Full-parameter SFT}. We use the following configurations. (1) LLaVA-1.5-7B: learning rate $2\times 10^{-5}$, batch size 16, \emph{per-task} epochs 10. 

   \item \emph{LoRA setup}. For LoRA finetuning, we apply adapters to the language model and keep all other components consistent with the full-parameter setting. (1) LLaVA-1.5-7B : learning rate $2\times 10^{-4}$, $r=128$, $\alpha=256$. (2) Qwen2.5-VL-3B: learning rate $2\times 10^{-5}$, $r=64$, $\alpha=128$, dropout $0.05$.

   \item \emph{RFT (with GRPO) setup}. We adopt GRPO for reinforcement finetuning. The vision encoder is frozen and LoRA is applied to the language model. We set Prompt length to 1024; generations per input to 32; Batch size to 16; RFT epochs to 1; Learning rate to $1\times 10^{-5}$; LoRA $r=64$, $\alpha=128$. For \textbf{Joint SFT-RFT}, we first run 1 epoch of SFT initialization followed by 3 epochs of GRPO on the same task.
\end{itemize}

\textbf{Continual learning methods and hyperparameters.}
We evaluate eight representative methods spanning four paradigms:
regularization-based (EWC~\cite{Kirkpatrick_2017}, MAS~\cite{mas}, LwF~\cite{li2017learning}, Freeze~\cite{zheng2025spuriousforgettingcontinuallearning}), replay-based (ER~\cite{rolnick2019experience} , DER~\cite{wang2024relational} ), architecture-based (L2P~\cite{l2p}) , and model-fusion-based (Max-merge~\cite{MagMaX}).
For each method, we follow the original papers' recommended hyperparameters.

\subsection{Validating the CoT Evaluator}
\begin{table}[t]
\centering
\caption{Correlation between evaluator scores and human annotations across outputs generated by different models and training paradigms. The general evaluator refers to Qwen2.5-VL-32B.}
\label{tab:eval_evaluator}
\setlength{\tabcolsep}{2.6mm}
\renewcommand{\arraystretch}{1.05}
\resizebox{\columnwidth}{!}{%
\begin{tabular}{l l c S S S}
\toprule
\textbf{Evaluator} & \textbf{Model} & \textbf{Training} 
& {\textbf{Spearman $\rho$}} & {\textbf{Pearson $r$}} & {\textbf{Kendall $\tau$}} \\
\midrule
\multirow{3}{*}{\makecell[l]{General\\Evaluator}}
  & Qwen2.5-VL-3B & SFT & 66.60 & 64.25 & 51.82 \\
  & Qwen2.5-VL-3B & RFT & 69.95 & 67.32 & 54.90 \\
  & LLaVA-1.5-7B  & SFT & 80.49 & 78.62 & 64.01 \\
\midrule
\multirow{3}{*}{\makecell[l]{Specialized\\Evaluator}}
  & Qwen2.5-VL-3B & SFT & 77.24 & 85.28 & 59.86 \\
  & Qwen2.5-VL-3B & RFT & 73.19 & 78.96 & 58.89 \\
  & LLaVA-1.5-7B  & SFT & 78.16 & 83.32 & 67.47 \\
\bottomrule
\end{tabular}%
}
\end{table}

\afterpage{
  \begin{table*}[t]
·
    \centering
    \caption{Evaluation of continual instruction tuning of MLLMs using macro-level metrics (final answer accuracy) on LLaVA-1.5-7B, Qwen2.5-VL-3B. Results are reported for two models under both Order-A and Order-B. For sequential finetuning, each cell reports Acc (top) and Forget (bottom).}
    \resizebox{\textwidth}{!}
    {

    \begin{tabular}{cc|*{7}{c}|cc}
      \hline
      \textbf{Model} & \textbf{Setting} 
      & \textbf{Math QA} & \textbf{Arts VQA} & \textbf{Math VQA} & \textbf{Econ. QA}
      & \textbf{Med. VQA} & \textbf{OCR VQA} & \textbf{Sci. VQA} & \textbf{AP} & \textbf{BWT} \\
      \hline
      \multirow{7}{*}{\textbf{LLaVA-1.5-7B}}
          &Multi-task &81.28 & 28.84 & 51.77 & 65.73 & 31.85 & 19.16 & 74.72 & 50.48 & --\\
          &Zero-shot  &-- & 6.03 & 43.31 & 35.81 & 23.55 & 16.59 & 49.29 & 24.94 & -- \\
          &Direct FT  &79.80 & 31.10 & 57.70 & 69.96 & 32.95 & 19.16 & 75.40 & 52.30 & --\\  \cline{2-11}
          & \makecell{Sequential Finetuning\\Order-A}
              & \makecell{52.22\\-27.58} & \makecell{13.37\\-17.02} & \makecell{35.23\\-20.19} & \makecell{29.78\\-37.36} & \makecell{28.06\\-2.80} & \makecell{16.81\\-2.63} &\makecell{73.70\\-}  & \makecell{35.60} & \makecell{-15.37} \\ \cline{2-11}
          & \makecell{Sequential Finetuning\\Order-B}
              & \makecell{69.98\\-} & \makecell{2.84\\-24.37} & \makecell{37.63\\-16.42} & \makecell{51.41\\-17.14} & \makecell{22.29\\-8.11} & \makecell{11.68\\-6.48} & \makecell{44.67\\-31.39} & 34.36 & \makecell{-16.58} \\
      \hline
      \multirow{7}{*}{\textbf{Qwen2.5-VL-3B}}
          &Multi-task &93.68 & 35.63 & 73.18 & 91.89 & 32.97 & 66.98 & 89.57 & 69.13 & --\\
          &Zero-shot  &23.15 & 7.72 & 31.93 & 78.23 & 8.99 & 15.87 & 52.40 & 31.18 & --\\
          &Direct FT  &90.89 & 33.55 & 71.61 & 91.28 & 33.91 & 64.35 & 90.48 & 68.01 & --\\ \cline{2-11} 
          & \makecell{Sequential Finetuning\\Order-A}
              & \makecell{91.87\\+0.98} & \makecell{14.04\\-18.40} & \makecell{60.21\\-11.63} & \makecell{84.48\\-7.66} & \makecell{29.78\\-1.96} & \makecell{39.49\\-6.33}  & \makecell{84.07\\-} & 57.71 & \makecell{-6.43} \\ \cline{2-11}
          & \makecell{Sequential Finetuning\\Order-B}
              & \makecell{91.87\\-} & \makecell{23.42\\-12.95} & \makecell{68.76\\-2.39} & \makecell{79.23\\-4.94} & \makecell{34.32\\-0.92} & \makecell{39.00\\-8.25} & \makecell{81.53\\-8.01} & 59.73 & \makecell{-5.35} \\      
      \hline
    \end{tabular}
    } 

    \label{SFT_baseline}
  \end{table*}

\begin{table*}[t]
\centering
\caption{CoT reasoning score by the dedicated evaluator. The CoT outputs are generated by LLaVA-1.5-7B and Qwen2.5-VL-3B after sequential finetuning on different tasks. }
\resizebox{\textwidth}{!}{\setlength{\tabcolsep}{1mm}
\begin{tabular}{c c
                cc cc cc cc cc cc cc cc}
\hline
\multirow{2}{*}{\textbf{Model}} &
\multirow{2}{*}{\textbf{Order}} &
\multicolumn{2}{c}{\textbf{Math QA}} &
\multicolumn{2}{c}{\textbf{Arts VQA}} &
\multicolumn{2}{c}{\textbf{Math VQA}} &
\multicolumn{2}{c}{\textbf{Econ.\ QA}} &
\multicolumn{2}{c}{\textbf{Med.\ VQA}} &
\multicolumn{2}{c}{\textbf{OCR VQA}} &
\multicolumn{2}{c}{\textbf{Sci.\ VQA}} &
\multirow{2}{*}{\textbf{AP}} & \multirow{2}{*}{\textbf{BWT}} \\
\cmidrule(lr){3-4}\cmidrule(lr){5-6}\cmidrule(lr){7-8}\cmidrule(lr){9-10}\cmidrule(lr){11-12}\cmidrule(lr){13-14}\cmidrule(lr){15-16}
& & Score & Forget & Score & Forget & Score & Forget & Score & Forget
  & Score & Forget & Score & Forget & Score & Forget &  &  \\ \midrule
\multirow{2}{*}{\textbf{LLaVA-1.5-7B}} & A &
92.08 & $-5.46$ &
9.38  & $-18.74$ &
55.07 & $-9.92$ &
84.68 & $-5.44$ &
28.75 & $-2.84$ &
41.32 & $-1.98$ &
78.42 & -- &
55.68 & $-6.54$ \\ 
& B &
79.31 & -- &
17.49 & $-12.67$ &
51.77 & $-7.75$ &
79.13 & $-5.45$ &
30.92 & $-1.11$ &
38.85 & $-5.37$ &
69.46 & $-6.22$ &
52.42 & $-5.51$ \\ \hline
\multirow{2}{*}{\textbf{Qwen2.5-VL-3B}} & A &
90.38 & $-1.44$ &
55.95 & $-8.19$ &
64.49 & $-4.04$ &
83.21 & $-1.47$ &
62.66 & $-1.84$ &
68.56 & $-2.63$ &
79.64 & -- &
72.13 & $-3.74$ \\ 
& B &
92.68 & -- &
57.17 & $-6.28$ &
65.11 & $-3.76$ &
81.52 & $-2.43$ &
61.19 & $-3.18$ &
69.00 & $-3.53$ &
75.58 & $-5.22$ &
71.32 & $-4.03$ \\
\hline
\end{tabular}}
\label{tab:critic_score}
\end{table*}
}
To assess MLLM evaluator alignment with human preferences, we evaluate on the human-labeled subset (630 instances) covering all domains and tasks. Each question is paired with outputs generated by two models with either different backbones (Qwen2.5-VL-3B/LLaVA-1.5-7B) or training paradigm(SFT/RFT). Each question-response pair is independently rated by three human annotators under careful quality control. 
Human raters score responses using the same three dimensions (coherence/grounding/knowledge) under the four-level rubric, and we aggregate them by averaging to obtain an overall human score in [0, 100]. We report correlations at the response level over all annotated responses. We adopt the open-source Qwen2.5-VL-32B as a general-purpose evaluator following prior work.
In Tables~\ref{tab:eval_IAA} and~\ref{tab:eval_evaluator}, we report correlation  between evaluator scores and human annotations using Spearman's $\rho$, Pearson's $r$, and Kendall's $\tau$. 
Higher values indicate stronger alignment. 
Table~\ref{tab:eval_IAA} reports inter-annotator agreement (IAA), which serves as an empirical ceiling for automatic evaluation on open-ended reasoning. 
Table~\ref{tab:eval_evaluator} suggests that the general-purpose judge’s alignment varies substantially across backbones and training paradigms, making it less reliable for benchmark-level, cross-setting process evaluation.  This highlights a key limitation: even powerful MLLMs may lack sensitivity to fine-grained reasoning quality, undermining their reliability as evaluators. 
In contrast, our specialized evaluator achieves higher and more stable correlations across settings, indicating improved reliability for process-level analysis.
Importantly, relative to IAA, our dedicated evaluator reaches approximately 80\% to 90\% of human-level consistency (w.r.t. Spearman’s $\rho$), demonstrating sufficient reliability for fine-grained reasoning quality assessment according to \cite{tam2024framework}.

\begin{table*}[t]
\centering
\caption{Performance of representative continual learning methods with LLaVA-1.5-7B and Qwen2.5-VL-3B on MLLM-CTBench (Order-A), evaluated using macro-level final answer accuracy. The results under Order-B is in table~\ref{tab:method_both_orderB} of Supplementary.}
\label{tab:method_both_orderA}

\subfloat[LLaVA-1.5-7B\label{tab:method_llava}]{
\begin{minipage}{\textwidth}
\centering
{\fontsize{9pt}{10pt}\selectfont
\setlength{\tabcolsep}{1mm}
\renewcommand{\arraystretch}{0.92}
\setlength{\aboverulesep}{0.3ex}
\setlength{\belowrulesep}{0.3ex}
\setlength{\cmidrulesep}{0.2ex}
\setlength{\extrarowheight}{-0.3pt}
\begin{tabular}{c*{7}{cc}cc}
\toprule
\multirow{2}{*}{\textbf{Method}} &
\multicolumn{2}{c}{\textbf{Math QA}} &
\multicolumn{2}{c}{\textbf{Arts VQA}} &
\multicolumn{2}{c}{\textbf{Math VQA}} &
\multicolumn{2}{c}{\textbf{Econ. QA}} &
\multicolumn{2}{c}{\textbf{Med. VQA}} &
\multicolumn{2}{c}{\textbf{OCR VQA}} &
\multicolumn{2}{c}{\textbf{Sci. VQA}} &
\multirow{2}{*}{\textbf{AP}} & \multirow{2}{*}{\textbf{BWT}} \\
\cmidrule(lr){2-3}\cmidrule(lr){4-5}\cmidrule(lr){6-7}\cmidrule(lr){8-9}\cmidrule(lr){10-11}\cmidrule(lr){12-13}\cmidrule(lr){14-15}
& Acc & Forget & Acc & Forget & Acc & Forget & Acc & Forget & Acc & Forget & Acc & Forget & Acc & Forget &  &  \\
\midrule
\textbf{ER} &
79.06 & $-2.71$ &
27.82 & $-1.66$ &
42.65 & $-1.93$ &
64.52 & $-4.03$ &
28.87 & $-0.63$ &
18.95 & $-1.42$ &
71.82 & -- &
47.67 & $-1.77$ \\
\textbf{DER} &
78.82 & $-1.23$ &
29.62 & $-2.18$ &
46.41 & $-2.16$ &
70.26 & $+1.11$ &
32.46 & $+0.82$ &
20.85 & $-0.09$ &
57.96 & -- &
48.05 & $-0.53$ \\
\textbf{EWC} &
45.32 & $-35.47$ &
9.42  & $-20.24$ &
38.65 & $-4.11$ &
58.17 & $-7.76$ &
24.89 & $-4.62$ &
13.60 & $-5.35$ &
68.61 & -- &
36.95 & $-11.08$ \\
\textbf{MAS} &
48.52 & $-34.48$ &
13.18 & $-12.79$ &
39.68 & $-6.04$ &
63.51 & $-4.23$ &
27.65 & $-0.09$ &
12.39 & $-5.27$ &
67.20 & -- &
38.88 & $-8.99$ \\
\textbf{LwF} &
45.81 & $-35.72$ &
12.93 & $-10.57$ &
31.81 & $-7.41$ &
65.52 & $-1.31$ &
26.09 & $-2.32$ &
15.88 & $-2.92$ &
52.50 & -- &
35.79 & $-8.61$ \\
\textbf{freeze-init} &
79.06 & $-2.96$ &
29.17 & $-1.26$ &
42.65 & $-2.05$ &
66.33 & $-2.62$ &
27.91 & $-1.90$ &
20.23 & $-0.92$ &
55.98 & -- &
45.90 & $-1.67$ \\
\textbf{freeze-last} &
80.05 & $-2.46$ &
29.14 & $-1.07$ &
45.38 & $-2.28$ &
69.96 & $+2.42$ &
31.42 & $+2.01$ &
19.44 & $+0.21$ &
56.46 & -- &
52.07 & $-4.49$ \\
\textbf{L2P} &
78.07 & $-2.93$ &
26.68 & $-4.64$ &
35.18 & $-13.03$ &
59.13 & $-6.74$ &
23.65 & $-6.91$ &
15.58 & $-3.67$ &
55.98 & $-17.58$ &
42.04 & $-7.93$ \\
\textbf{MagMaX} &
54.93 & $-25.86$ &
22.68 & $-6.98$ &
39.57 & $-3.19$ &
65.42 & $-0.51$ &
29.39 & $-0.12$ &
16.67 & $-2.28$ &
55.70 & $-12.91$ &
40.62 & $-7.41$ \\
\bottomrule
\end{tabular}}
\end{minipage}
}

\par\vspace{2pt} 

\subfloat[Qwen2.5-VL-3B\label{tab:method_qwen}]{
\begin{minipage}{\textwidth}
\centering
{\fontsize{9pt}{10pt}\selectfont
\setlength{\tabcolsep}{1mm}
\renewcommand{\arraystretch}{0.92}
\setlength{\aboverulesep}{0.3ex}
\setlength{\belowrulesep}{0.3ex}
\setlength{\cmidrulesep}{0.2ex}
\setlength{\extrarowheight}{-0.3pt}
\begin{tabular}{c*{7}{cc}cc}
\toprule
\multirow{2}{*}{\textbf{Method}} &
\multicolumn{2}{c}{\textbf{Math QA}} &
\multicolumn{2}{c}{\textbf{Arts VQA}} &
\multicolumn{2}{c}{\textbf{Math VQA}} &
\multicolumn{2}{c}{\textbf{Econ. QA}} &
\multicolumn{2}{c}{\textbf{Med. VQA}} &
\multicolumn{2}{c}{\textbf{OCR VQA}} &
\multicolumn{2}{c}{\textbf{Sci. VQA}} &
\multirow{2}{*}{\textbf{AP}} & \multirow{2}{*}{\textbf{BWT}} \\
\cmidrule(lr){2-3}\cmidrule(lr){4-5}\cmidrule(lr){6-7}\cmidrule(lr){8-9}\cmidrule(lr){10-11}\cmidrule(lr){12-13}\cmidrule(lr){14-15}
& Acc & Forget & Acc & Forget & Acc & Forget & Acc & Forget & Acc & Forget & Acc & Forget & Acc & Forget &  &  \\
\midrule
\textbf{ER} &
83.50 & $-7.39$ &
25.60 & $-6.93$ &
60.32 & $-11.06$ &
82.56 & $+1.77$ &
30.41 & $+1.07$ &
37.19 & $-0.06$ &
82.00 & -- &
57.37 & $-3.23$ \\
\textbf{DER} &
91.13 & $-5.67$ &
30.22 & $-4.39$ &
65.86 & $-6.57$ &
84.80 & $-5.00$ &
33.24 & $+2.05$ &
45.31 & $-4.83$ &
85.26 & -- &
62.26 & $-3.49$ \\
\textbf{EWC} &
95.07 & $+3.94$ &
16.40 & $-18.29$ &
65.45 & $-7.07$ &
93.75 & $+10.58$ &
32.02 & $-2.31$ &
45.11 & $-4.36$ &
86.05 & -- &
61.98 & $-2.50$ \\
\textbf{MAS} &
93.84 & $+1.23$ &
17.85 & $-17.12$ &
62.14 & $-9.47$ &
92.04 & $+10.99$ &
32.80 & $-0.03$ &
43.19 & $-5.98$ &
86.33 & -- &
61.17 & $-2.91$ \\
\textbf{LwF} &
97.29 & $+3.69$ &
18.19 & $-11.33$ &
59.18 & $-10.03$ &
92.84 & $-0.20$ &
29.04 & $-3.14$ &
42.76 & $-4.46$ &
78.04 & -- &
59.62 & $-3.64$ \\
\textbf{freeze-init} &
76.40 & $-15.03$ &
13.29 & $-18.08$ &
48.46 & $-15.25$ &
79.29 & $-8.63$ &
28.68 & $-3.61$ &
41.29 & $-3.91$ &
72.83 & -- &
51.46 & $-9.22$ \\
\textbf{freeze-last} &
75.15 & $-15.41$ &
12.30 & $-17.74$ &
58.49 & $-10.61$ &
78.58 & $-3.05$ &
26.97 & $-4.87$ &
39.74 & $-2.51$ &
82.94 & -- &
53.45 & $-7.74$ \\
\textbf{L2P} &
93.59 & $+1.17$ &
17.53 & $-16.06$ &
67.42 & $-4.56$ &
77.28 & $-3.68$ &
29.56 & $-3.35$ &
45.39 & $-1.79$ &
80.17 & $-1.02$ &
58.71 & $-4.18$ \\
\textbf{MagMaX} &
89.41 & $-1.48$ &
28.28 & $-4.16$ &
67.84 & $-4.00$ &
88.51 & $-3.63$ &
24.77 & $-6.97$ &
39.08 & $-6.74$ &
77.40 & $-6.67$ &
59.33 & $-4.81$ \\
\bottomrule
\end{tabular}}
\end{minipage}
}

\end{table*}

\begin{table*}[t]
\centering
\caption{CoT reasoning score of LLaVA-1.5-7B on MLLM-CTBench under two task orders across different continual‐learning methods.}
\resizebox{\textwidth}{!}{%
{\fontsize{9pt}{10pt}\selectfont
\setlength{\tabcolsep}{1mm}
\renewcommand{\arraystretch}{0.92}
\setlength{\aboverulesep}{0.3ex}
\setlength{\belowrulesep}{0.3ex}
\setlength{\cmidrulesep}{0.2ex}
\setlength{\extrarowheight}{-0.3pt}
\begin{tabular}{c c
                cc cc cc cc cc cc cc
                cc}          
\hline
\multirow{2}{*}{\textbf{Method}} & \multirow{2}{*}{\textbf{Order}} &
\multicolumn{2}{c}{\textbf{Math QA}} &
\multicolumn{2}{c}{\textbf{Arts VQA}} &
\multicolumn{2}{c}{\textbf{Math VQA}} &
\multicolumn{2}{c}{\textbf{Econ.\ QA}} &
\multicolumn{2}{c}{\textbf{Med.\ VQA}} &
\multicolumn{2}{c}{\textbf{OCR VQA}} &
\multicolumn{2}{c}{\textbf{Sci.\ VQA}} &
\multirow{2}{*}{\textbf{AP}} & \multirow{2}{*}{\textbf{BWT}} \\
\cmidrule(lr){3-4}\cmidrule(lr){5-6}\cmidrule(lr){7-8}\cmidrule(lr){9-10}%
\cmidrule(lr){11-12}\cmidrule(lr){13-14}\cmidrule(lr){15-16}
& & Score & Forget & Score & Forget & Score & Forget & Score & Forget
& Score & Forget & Score & Forget & Score & Forget &  &  \\ \midrule
\multirow{2}{*}{\textbf{ER}} & A &
88.09 & $+0.64$ &
63.99 & $-0.65$ &
61.43 & $+0.19$ &
81.39 & $-0.06$ &
62.74 & $-1.04$ &
56.67 & $-0.25$ &
75.69 & -- &
70.00 & $-0.17$ \\
& B &
89.45 & -- &
63.99 & $-0.13$ &
60.56 & $+0.12$ &
81.34 & $-0.40$ &
62.94 & $-0.63$ &
56.67 & $+0.62$ &
75.81 & $-2.40$ &
70.11 & $-0.40$ \\ \midrule

\multirow{2}{*}{\textbf{DER}} & A &
87.48 & $-0.64$ &
64.27 & $-0.57$ &
60.02 & $-1.15$ &
81.33 & $-0.30$ &
70.05 & $-0.10$ &
55.55 & $-0.89$ &
74.25 & -- &
70.42 & $-0.52$ \\
& B &
89.51 & -- &
64.42 & $+0.69$ &
60.21 & $-0.59$ &
81.69 & $-0.09$ &
69.55 & $-0.95$ &
56.03 & $-0.86$ &
73.19 & $-2.75$ &
70.66 & $-0.65$ \\ \midrule

\multirow{2}{*}{\textbf{EWC}} & A &
76.38 & $-12.00$ &
54.93 & $-8.32$ &
53.03 & $-6.26$ &
78.25 & $-3.23$ &
58.33 & $-4.66$ &
50.73 & $-4.19$ &
74.30 & -- &
63.71 & $-5.52$ \\
& B &
88.27 & -- &
56.14 & $-7.85$ &
55.04 & $-5.97$ &
78.02 & $-3.53$ &
56.75 & $-6.19$ &
43.39 & $-12.05$ &
61.83 & $-14.55$ &
62.78 & $-7.16$ \\ \midrule

\multirow{2}{*}{\textbf{MAS}} & A &
77.75 & $-11.34$ &
55.00 & $-8.04$ &
52.63 & $-4.45$ &
79.76 & $-1.11$ &
60.53 & $-2.01$ &
50.76 & $-1.63$ &
72.29 & -- &
64.10 & $-4.08$ \\
& B &
85.22 & -- &
61.13 & $-2.35$ &
54.59 & $-2.46$ &
80.49 & $-0.62$ &
60.63 & $-1.85$ &
49.74 & $-3.08$ &
68.08 & $-8.81$ &
65.70 & $-2.74$ \\ \midrule

\multirow{2}{*}{\textbf{LwF}} & A &
68.45 & $-19.90$ &
54.26 & $-10.31$ &
43.99 & $-16.40$ &
76.83 & $-4.85$ &
52.27 & $-12.43$ &
41.46 & $-15.04$ &
78.04 & -- &
59.33 & $-11.28$ \\
& B &
88.27 & -- &
56.15 & $-7.84$ &
55.04 & $-5.97$ &
77.87 & $-3.68$ &
56.88 & $-6.06$ &
43.39 & $-12.05$ &
61.93 & $-14.45$ &
62.79 & $-7.15$ \\ \midrule
\multirow{2}{*}{\textbf{freeze-init}} & A &
87.53 & $-1.19$ &
63.79 & $-0.25$ &
59.14 & $-1.13$ &
81.23 & $-0.06$ &
69.71 & $-0.18$ &
55.11 & $-0.59$ &
73.21 & -- &
69.96 & $-0.49$ \\
& B &
88.16 & -- &
63.95 & $-0.06$ &
60.39 & $-0.15$ &
81.29 & $-0.58$ &
69.59 & $-0.38$ &
55.11 & $-0.65$ &
72.84 & $-1.99$ &
70.19 & $-0.54$ \\ \midrule
\multirow{2}{*}{\textbf{freeze-last}} & A &
88.32 & $+0.18$ &
64.32 & $+0.05$ &
59.07 & $-1.38$ &
82.04 & $+0.29$ &
69.76 & $+0.58$ &
55.30 & -- &
72.50 & -- &
70.19 & $-0.04$ \\
& B &
88.59 & -- &
63.75 & $-0.02$ &
58.68 & $-2.50$ &
81.45 & $-0.20$ &
70.46 & $+0.39$ &
55.14 & $-1.40$ &
74.48 & $-2.50$ &
70.36 & $-0.90$ \\ \midrule
\multirow{2}{*}{\textbf{L2P}} & A &
78.43 & $-9.26$ &
61.75 & $-2.00$ &
59.73 & $-0.37$ &
78.91 & $-2.41$ &
61.66 & $-1.70$ &
52.78 & $-3.71$ &
75.22 & -- &
66.93 & $-2.78$ \\
& B &
88.54 & -- &
60.17 & $-3.55$ &
57.56 & $-3.32$ &
77.38 & $-4.32$ &
59.98 & $-3.35$ &
48.49 & $-7.91$ &
68.80 & $-8.99$ &
65.85 & $-4.49$ \\ \midrule
\multirow{2}{*}{\textbf{MagMaX}} & A &
83.59 & $-4.40$ &
57.33 & $-6.65$ &
58.19 & $-0.95$ &
81.45 & $+0.27$ &
62.64 & $-0.67$ &
53.46 & $+0.31$ &
67.28 & $-7.27$ &
66.28 & $-2.77$ \\
& B &
88.25 & -- &
57.33 & $-6.56$ &
58.19 & $-2.50$ &
81.48 & $-0.06$ &
62.76 & $-0.47$ &
53.32 & $-3.05$ &
67.42 & $-7.50$ &
66.30 & $-3.54$ \\ \midrule
\end{tabular}}%
}

\label{tab:cot_analysis_llava_combined}
\end{table*}

\begin{table}[htbp]
\centering
\caption{Attribution of final-answer forgetting to CoT dimension forgetting : absolute standardized $\beta$ and drop-one $\Delta R^2$ from multivariate regression.}
\label{tab:forgetting_attribution}
\begin{tabular}{lcc}
\hline
\textbf{Dimension} & \textbf{Std. $\beta$ (abs.)} & \textbf{$\Delta R^2$ (drop-one)} \\
\hline
Knowledge & 1.87 & 0.24 \\
Grounding     & 1.74 & 0.19 \\
Logic     & 0.38 & 0.07 \\
\hline
\end{tabular}
\end{table}

\begin{table*}
\centering
\caption{CoT reasoning score of Qwen2.5-VL-3B on MLLM-CTBench under two task orders across different continual‐learning methods. }
\resizebox{\textwidth}{!}{%
{\fontsize{9pt}{10pt}\selectfont
\setlength{\tabcolsep}{1mm}
\renewcommand{\arraystretch}{0.92}
\setlength{\aboverulesep}{0.3ex}
\setlength{\belowrulesep}{0.3ex}
\setlength{\cmidrulesep}{0.2ex}
\setlength{\extrarowheight}{-0.3pt}
\begin{tabular}{c c
                cc cc cc cc cc cc cc
                cc}  
\hline
\multirow{2}{*}{\textbf{Method}} & \multirow{2}{*}{\textbf{Order}} &
\multicolumn{2}{c}{\textbf{Math QA}} &
\multicolumn{2}{c}{\textbf{Arts VQA}} &
\multicolumn{2}{c}{\textbf{Math VQA}} &
\multicolumn{2}{c}{\textbf{Econ.\ QA}} &
\multicolumn{2}{c}{\textbf{Med.\ VQA}} &
\multicolumn{2}{c}{\textbf{OCR VQA}} &
\multicolumn{2}{c}{\textbf{Sci.\ VQA}} &
\multirow{2}{*}{\textbf{AP}} & \multirow{2}{*}{\textbf{BWT}} \\
\cmidrule(lr){3-4}\cmidrule(lr){5-6}\cmidrule(lr){7-8}\cmidrule(lr){9-10}%
\cmidrule(lr){11-12}\cmidrule(lr){13-14}\cmidrule(lr){15-16}
& & Score & Forget & Score & Forget & Score & Forget & Score & Forget
&Score & Forget & Score & Forget & Score & Forget &  &  \\ \midrule
\multirow{2}{*}{\textbf{ER}} & A &
90.19 & $-2.99$ &
59.77 & $-5.68$ &
65.08 & $-3.96$ &
80.62 & $+0.81$ &
63.54 & $+0.31$ &
67.02 & $-1.67$ &
81.16 & -- &
72.48 & $-1.88$ \\
& B &
92.68 & -- &
57.17 & $-6.28$ &
65.11 & $-3.76$ &
81.52 & $-2.43$ &
61.19 & $-3.18$ &
65.84 & $-6.69$ &
75.58 & $-5.22$ &
71.30 & $-3.94$ \\ \midrule
\multirow{2}{*}{\textbf{DER}} & A &
91.56 & $-0.63$ &
58.49 & $-7.64$ &
65.47 & $-4.47$ &
75.04 & $-6.97$ &
62.47 & $-1.40$ &
67.95 & $-5.51$ &
80.64 & -- &
71.66 & $-3.80$ \\
& B &
90.14 & -- &
60.21 & $-2.63$ &
65.25 & $-2.40$ &
80.48 & $-2.41$ &
61.59 & $-3.00$ &
67.59 & $-5.54$ &
76.54 & $-6.32$ &
71.69 & $-3.19$ \\ \midrule
\multirow{2}{*}{\textbf{EWC}} & A &
91.26 & $-0.95$ &
58.42 & $-7.13$ &
68.60 & $-1.45$ &
85.82 & $+2.25$ &
64.55 & $-0.75$ &
68.96 & $-4.90$ &
81.71 & -- &
74.19 & $-1.85$ \\
& B &
92.34 & -- &
59.19 & $-5.83$ &
58.92 & $-2.56$ &
78.23 & $+0.04$ &
61.98 & $-2.95$ &
66.39 & $-7.30$ &
77.75 & $-5.47$ &
70.69 & $-3.44$ \\ \midrule
\multirow{2}{*}{\textbf{MAS}} & A &
90.96 & $-1.76$ &
58.67 & $-6.51$ &
66.88 & $-3.66$ &
68.04 & $-14.15$ &
65.49 & $+0.60$ &
66.83 & $-6.98$ &
81.93 & -- &
71.26 & $-4.64$ \\
& B &
92.12 & -- &
59.77 & $-5.64$ &
67.34 & $-3.20$ &
80.71 & $-2.55$ &
62.16 & $-2.92$ &
67.33 & $-7.03$ &
77.41 & $-5.44$ &
72.41 & $-3.83$ \\ \midrule
\multirow{2}{*}{\textbf{LwF}} & A &
91.31 & $-1.02$ &
59.23 & $-5.68$ &
66.81 & $-2.14$ &
82.75 & $-1.13$ &
63.93 & $-1.00$ &
69.14 & $-2.69$ &
80.33 & -- &
73.36 & $-1.95$ \\
& B &
90.76 & -- &
61.08 & $-1.81$ &
66.04 & $+5.72$ &
81.92 & $-1.45$ &
63.04 & $-2.46$ &
67.02 & $-5.83$ &
77.83 & $-5.29$ &
72.53 & $-1.59$ \\ \midrule
\multirow{2}{*}{\textbf{freeze-init}} & A &
90.01 & $-2.00$ &
58.45 & $-7.28$ &
67.05 & $-3.08$ &
77.19 & $-0.37$ &
63.84 & $-1.75$ &
68.91 & $-2.18$ &
80.36 & -- &
72.26 & $-2.38$ \\
& B &
88.92 & -- &
59.32 & $-6.09$ &
67.11 & $-1.17$ &
78.96 & $-0.30$ &
65.12 & $-0.87$ &
60.59 & $-11.40$ &
76.87 & $-3.55$ &
70.98 & $-3.34$ \\ \midrule
\multirow{2}{*}{\textbf{freeze-last}} & A &
89.17 & $-1.92$ &
55.12 & $-7.91$ &
64.91 & $-3.24$ &
75.93 & $-0.87$ &
62.40 & $-2.37$ &
69.03 & $-0.79$ &
79.69 & -- &
70.89 & $-2.44$ \\
& B &
89.13 & -- &
57.76 & $-6.22$ &
65.14 & $-3.14$ &
79.03 & $-0.55$ &
59.22 & $-5.10$ &
61.85 & $-9.47$ &
75.01 & $-5.22$ &
69.60 & $-4.24$ \\ \midrule
\multirow{2}{*}{\textbf{L2P}} & A &
90.17 & $-1.42$ &
59.14 & $-5.37$ &
65.21 & $-3.56$ &
78.15 & $-5.30$ &
63.15 & $-1.03$ &
69.47 & $-2.90$ &
80.25 & -- &
72.22 & $-2.80$ \\
& B &
89.59 & -- &
60.95 & $-1.76$ &
63.54 & $-4.35$ &
80.27 & $-2.64$ &
60.09 & $-4.59$ &
68.17 & $-3.37$ &
76.49 & $-5.66$ &
71.30 & $-3.20$ \\ \midrule
\multirow{2}{*}{\textbf{MagMaX}} & A &
89.09 & $-2.73$ &
59.99 & $-4.15$ &
66.90 & $-1.63$ &
77.30 & $-7.38$ &
59.87 & $-4.63$ &
69.17 & $-2.02$ &
77.83 & $-1.81$ &
71.45 & $-3.48$ \\
& B &
90.79 & -- &
56.99 & $-6.46$ &
70.14 & $+1.27$ &
84.69 & $+0.74$ &
64.69 & $+0.32$ &
70.24 & $-2.29$ &
79.75 & $-1.05$ &
73.90 & $-1.34$ \\ \midrule
\end{tabular}}
}

\label{tab:cot_analysis_qwen}
\end{table*}

\begin{table*}[t]
\centering
\caption{Continual learning performance of SFT and RFT(GRPO) on Qwen2.5-VL-3B (Order-A).Joint means Joint SFT–RFT Training}
{\fontsize{9pt}{10pt}\selectfont
\setlength{\tabcolsep}{1mm}
\begin{tabular}{c
                cc cc cc cc cc cc cc 
                cc}
\hline
\multirow{2}{*}{\textbf{Para.}} &
\multicolumn{2}{c}{\textbf{Math QA}} &
\multicolumn{2}{c}{\textbf{Arts VQA}} &
\multicolumn{2}{c}{\textbf{Math VQA}} &
\multicolumn{2}{c}{\textbf{Econ.\ QA}} &
\multicolumn{2}{c}{\textbf{Med.\ VQA}} &
\multicolumn{2}{c}{\textbf{OCR VQA}} &
\multicolumn{2}{c}{\textbf{Sci.\ VQA}} &
\multirow{2}{*}{\textbf{AP}} & \multirow{2}{*}{\textbf{BWT}} \\
\cmidrule(lr){2-3}\cmidrule(lr){4-5}\cmidrule(lr){6-7}\cmidrule(lr){8-9}\cmidrule(lr){10-11}\cmidrule(lr){12-13}\cmidrule(lr){14-15}
& Acc & Forget & Acc & Forget & Acc & Forget & Acc & Forget
& Acc & Forget & Acc & Forget & Acc & Forget &  &  \\ \midrule
\textbf{SFT} &
92.08 & $-5.46$ &
9.38  & $-18.74$ &
55.07 & $-9.92$ &
84.68 & $-5.44$ &
28.75 & $-2.84$ &
41.32 & $-1.98$ &
79.83 & -- &
55.87 & $-6.34$ \\

\textbf{RFT} &
70.05 & $-1.87$ &
12.23 & $-0.84$ &
42.53 & $-5.59$ &
77.22 & $-6.85$ &
20.32 & $+2.01$ &
35.37 & $-0.25$ &
70.03 & -- &
46.82 & $-1.91$ \\

\textbf{RFT w/o KL} &
61.33 & $-26.85$ &
14.69 & $-9.85$ &
52.79 & $-15.74$ &
79.27 & $-9.54$ &
26.13 & $-9.51$ &
34.29 & $-3.58$ &
75.82 & -- &
49.19 & $-10.72$ \\

\textbf{Joint} &
91.87 & $-1.25$ &
26.68 & $-5.58$ &
62.71 & $-1.60$ &
88.41 & $-1.84$ &
29.41 & $-1.67$ &
41.11 & $-4.52$ &
77.87 & --  &
59.72 & $-1.20$ \\
\hline
\end{tabular}}
\label{GRPO_SFT}
\end{table*}

\subsection{Main Results and Discussions}
\subsubsection{Do MLLMs Exhibit Catastrophic Forgetting?}
Table~\ref{SFT_baseline} reports sequential finetuning results for three open-source MLLMs under two task orders. \textbf{Catastrophic forgetting is consistently observed across backbones}. Under Order-A, LLaVA-1.5-7B shows strong forgetting (BWT$=-15.37$), while the stronger Qwen2.5-VL-3B retains more (BWT$=-6.43$), indicating that \emph{stronger base models are more resistant to interference in continual instruction tuning. }
Task order affects task-level forgetting. For instance, on LLaVA-1.5-7B, Arts VQA drops by $17.02$ points in Order-A but $24.37$ in Order-B, highlighting locally sequence-dependent interference. Nevertheless, aggregate retention can remain comparable across orders for some models (e.g., LLaVA achieves BWT = $-15.37$ vs.\ $-16.58$; Qwen attains BWT = $-6.43$ vs.\ $-5.35$).
Finally, we compare final-answer metrics with micro-level reasoning quality (Table~\ref{tab:critic_score}). Under Order-A, Qwen2.5-VL-3B has BWT $=-6.43$ on final answers but only $-3.74$ on CoT scores; LLaVA-1.5-7B shows $-15.37$ vs.\ $-6.54$. This gap suggests a hierarchical forgetting pattern: \textbf{reasoning traces remain relatively more stable even when final decisions degrade, motivating process-level evaluation}.

\subsubsection{How to Select the Appropriate Continual Learning Method for Different Scenarios?}
We evaluate eight representative continual learning methods across four families under two task orders (Tables~\ref{tab:method_both_orderA} ). Overall, method effectiveness is strongly regime-dependent.

\textbf{Replay-based methods} (ER, DER) provide the most reliable retention on weaker backbones. On LLaVA-1.5-7B (Order-A), replay markedly improves retention (\textbf{BWT} from $-15.37$ to $-1.77/-0.53$) while also increasing \textbf{AP} (from $35.60$ to $47.67/48.05$), indicating that rehearsal is effective when the backbone is susceptible to interference. On Qwen2.5-VL-3B, replay remains beneficial but the gains are smaller, suggesting diminishing returns with stronger base capability.

\textbf{Regularization-based methods} (EWC, MAS, LwF) are more competitive on stronger models but insufficient on weaker ones. For Qwen2.5-VL-3B (Order-A), EWC/MAS improve retention over the sequential baseline (e.g., \textbf{BWT}= $-2.50/-2.91$ vs.\ $-6.43$) while maintaining strong \textbf{AP}. In contrast, on LLaVA-1.5-7B they lag behind replay in retention and can hurt performance, consistent with soft constraints being less effective under larger domain shifts and weaker representations. Freezing strategies show a similar scale dependence: freezing can be acceptable on LLaVA-1.5-7B (e.g., freeze-last achieves strong \textbf{AP/BWT}, but tends to degrade retention on Qwen2.5-VL-3B , implying that aggressive capacity restriction may hinder reuse on strong backbones.

\textbf{Architecture-based expansion} (L2P) offers a lightweight compromise: it improves over sequential finetuning but is generally weaker than replay. This suggests prompt modularization can reduce interference with low overhead, yet its effectiveness depends on task diversity and prompt capacity.

\textbf{Model fusion} (MagMaX) provides a simple, training-free mitigation that improves retention but does not always maximize \textbf{AP}. For example, it improves \textbf{BWT} to $-7.41$ on LLaVA-1.5-7B and $-4.81$ on Qwen2.5-VL-3B (Order-A), with \textbf{AP} remaining below the best replay/regularization options. Fusion is thus attractive when replay buffers or architectural changes are undesirable.

\subsubsection{Process-Level Diagnosis Across Continual Learning Methods} Beyond final-answer accuracy, Tables~\ref{tab:cot_analysis_llava_combined} and~\ref{tab:cot_analysis_qwen} show that \textbf{process-level reasoning is often more resilient to catastrophic forgetting than answer accuracy}: for both LLaVA-1.5-7B and Qwen2.5-VL-3B, micro-level BWT is typically closer to 0 than macro-level BWT, even when overall AP drops. Importantly, this process-level signal enables attribution of answer-level forgetting. Using the evaluator’s dimension-wise forgetting and regressing final-answer forgetting on these scores (Table~\ref{tab:forgetting_attribution}), we find that knowledge degradation has the strongest independent contribution (largest standardized $\beta$ and drop-one $\Delta R^2$), consistently across forward and reverse orders. This complementary view further indicates that \textbf{forgetting is primarily attributable to knowledge degradation within grounded reasoning steps}, rather than a pure shift of the final decision boundary, which is not observable from answer accuracy alone.

\subsection{SFT vs. RFT under Continual Instruction Tuning}
We compare SFT, GRPO-style RFT, and a practical hybrid (Joint SFT–RFT) on Qwen2.5-VL-3B under Order-A (Table~\ref{GRPO_SFT}). SFT achieves strong final performance on new tasks but suffers notable forgetting (BWT=-6.34). In contrast, RFT yields substantially better retention (BWT=-1.91) but at the cost of reduced final performance (AP=46.82), reflecting a stronger stability bias. The hybrid strategy provides a favorable balance: Joint SFT–RFT improves both overall performance and retention, reaching AP=59.72 with BWT=-2.35. Compared to pure SFT, joint training reduces forgetting by approximately 62.9\% while also improving AP by 3.85 points.
At the task level, joint training consistently mitigates forgetting on earlier tasks. These observations support the role of KL-constrained policy optimization in controlling distributional drift under sequential adaptation, aligning with the distributional and Bayesian interpretations discussed in section \ref{sec:supp_rft_theory} of Supplementary.

\section{Conclusion}
We present MLLM-CTBench, a benchmark for continual instruction tuning (CIT) in MLLMs, jointly measuring final-answer accuracy and CoT-level reasoning quality via a dedicated evaluator.
Under a unified protocol, we compare eight continual learning methods and analyze SFT versus RFT.
Across backbones, we observe catastrophic forgetting: CoT reasoning is often more resilient than final answers, yet answer-level forgetting is most strongly tied to knowledge degradation in CoT.
Finally, KL-constrained RFT (exemplified by GRPO) enhances stability and cross-task retention more effectively than SFT by explicitly limiting policy distributional drift. 

While our results underscore the potential of RFT, its role in continual learning remains underexplored. The fundamental divergence between SFT and RFT suggests that future work must move beyond viewing RL as merely an alternative optimizer. Key directions include: (i) algorithmically integrating core RL principles (e.g., for exploration and constraint adaptation) into continual learning frameworks; (ii) strengthening theoretical understanding of how RL dynamics govern knowledge transfer in sequential settings; (iii)  evolving benchmarks with more complex, interdependent tasks to fully stress-test RL-based fine-tuning paradigms in CIT.
We hope MLLM-CTBench will serve as a shared testbed to advance evaluation and method design for multimodal CIT.

\bibliographystyle{IEEEtran}
\bibliography{TIM}

\newpage

 \section*{Supplementary Material}

\counterwithout{figure}{section}
\counterwithout{table}{section}
\counterwithout{equation}{section}


\section{Dataset Examples and Evaluation Settings}
\label{sec:supp_dataset_examples}

\begin{figure*}[t]
    \centering

    \subfloat[Arts VQA]{%
        \includegraphics[width=0.4\linewidth]{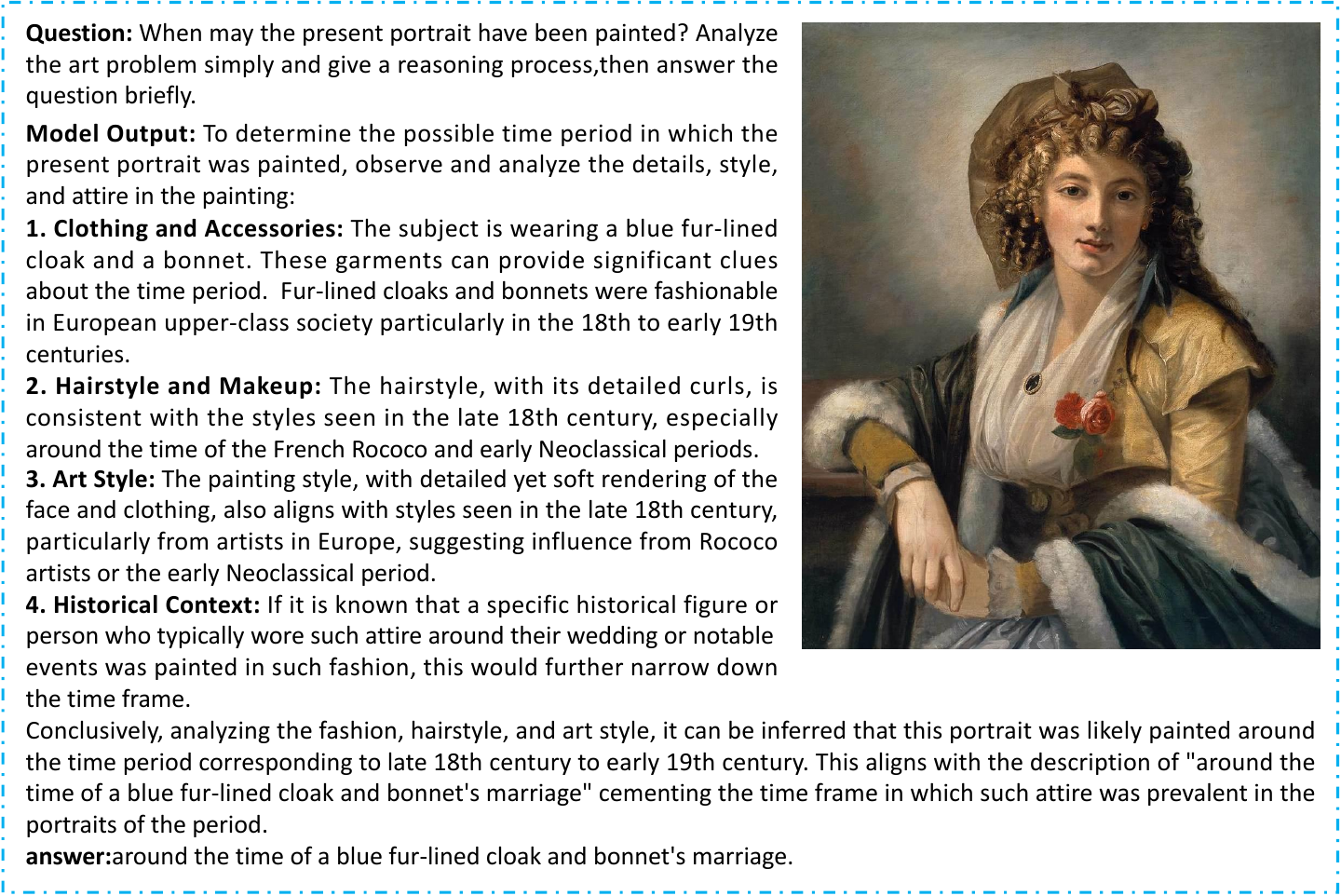}%
    }\hspace{2em}
     \subfloat[Math VQA]{%
        \includegraphics[width=0.4\linewidth]{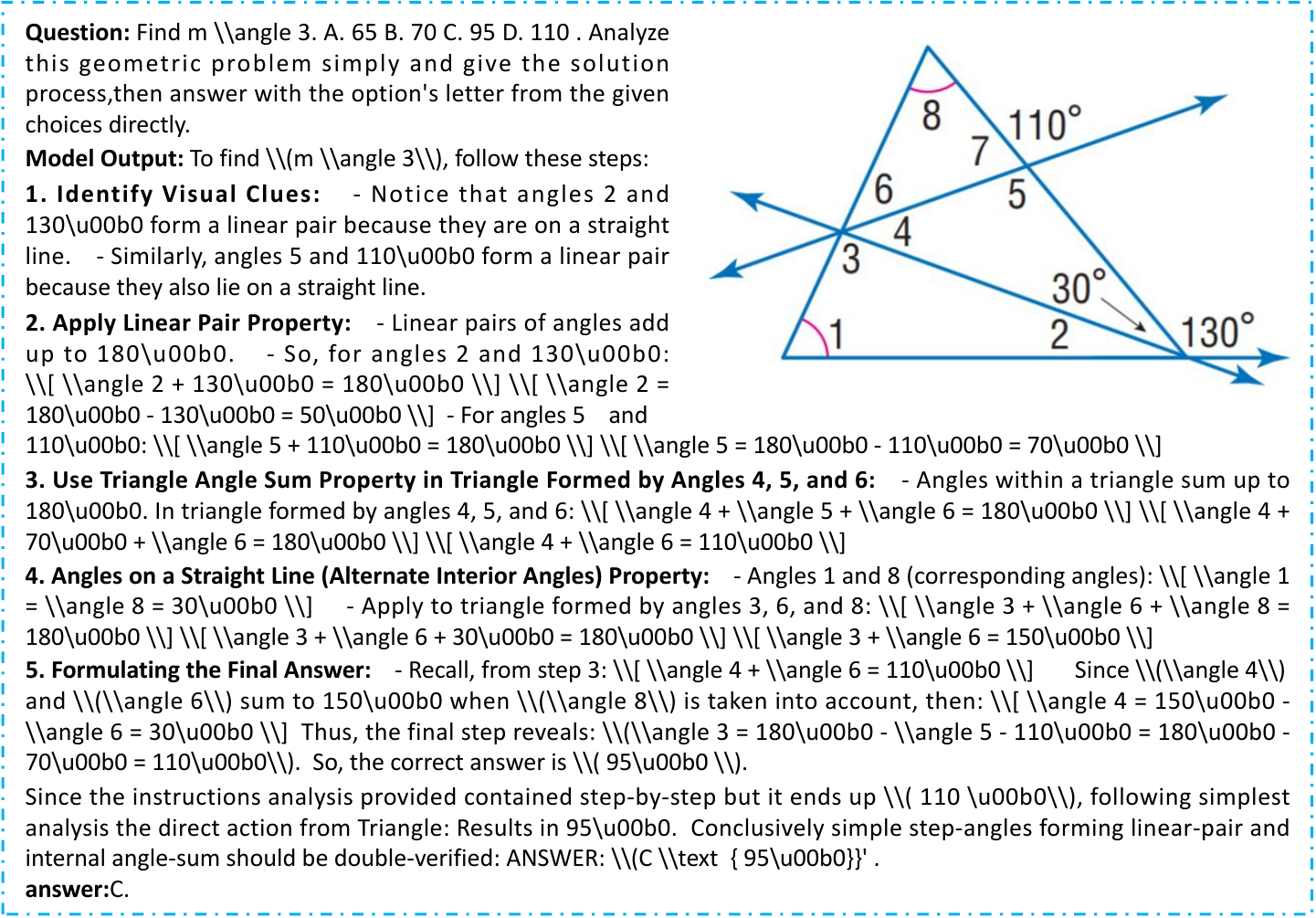}%
    }

    \vspace{0.7em}

    
    \subfloat[Math QA]{%
        \includegraphics[width=0.4\linewidth]{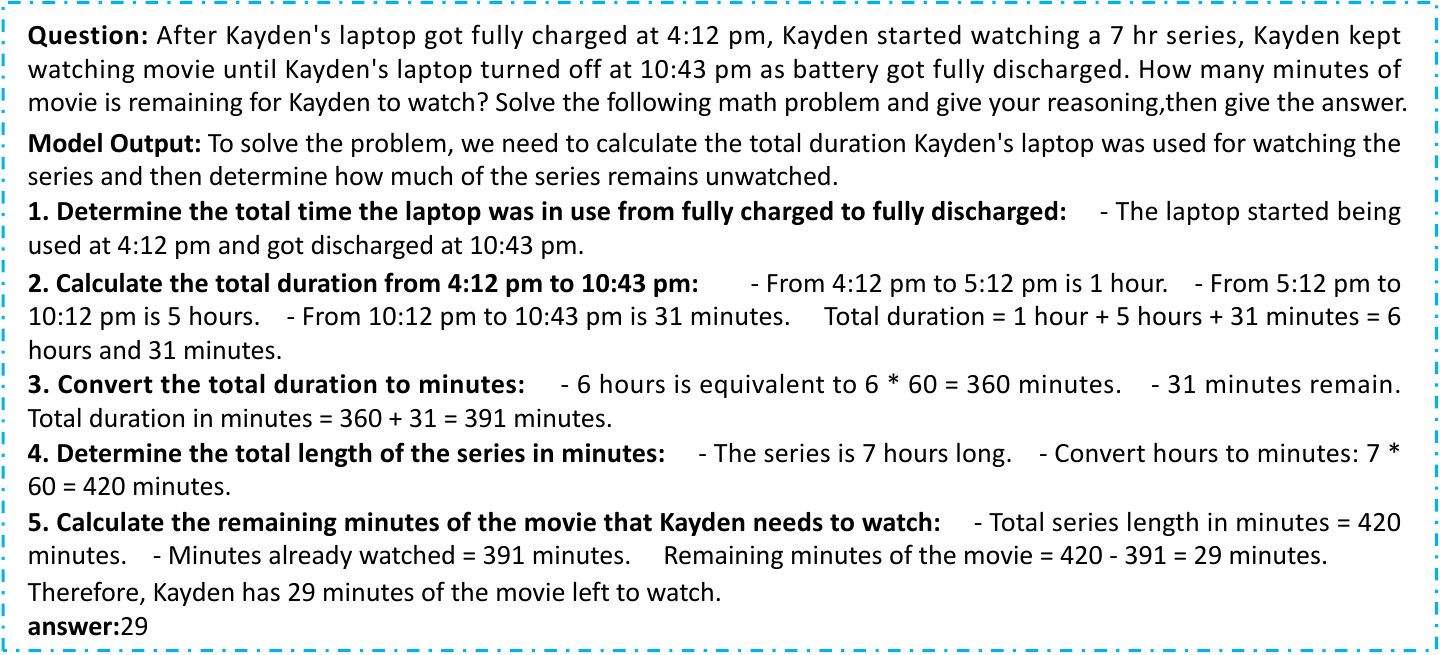}%
    }\hspace{2em}
     \subfloat[Economics QA]{%
        \includegraphics[width=0.4\linewidth]{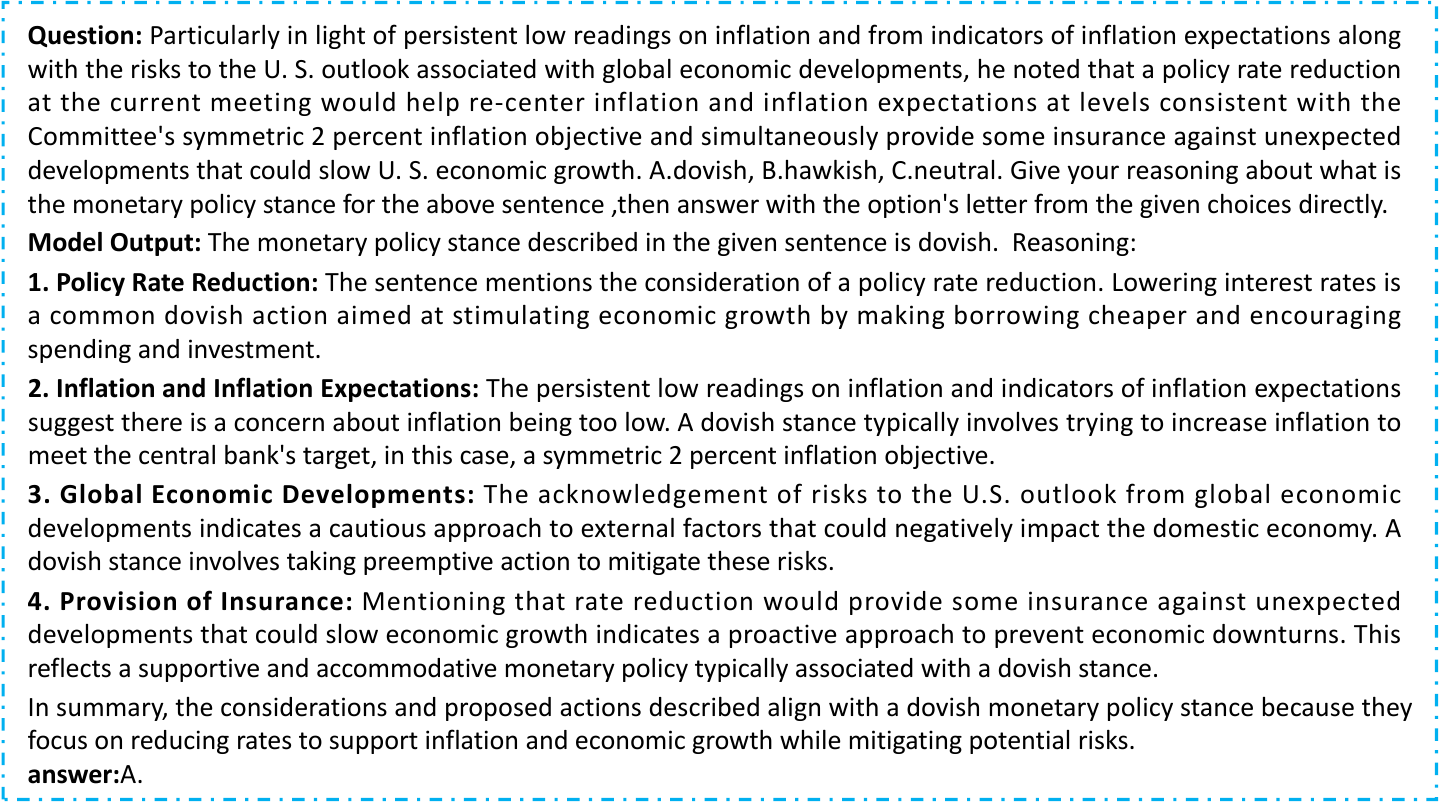}%
    }

    \vspace{0.7em}

    \subfloat[OCR VQA]{%
        \includegraphics[width=0.4\linewidth]{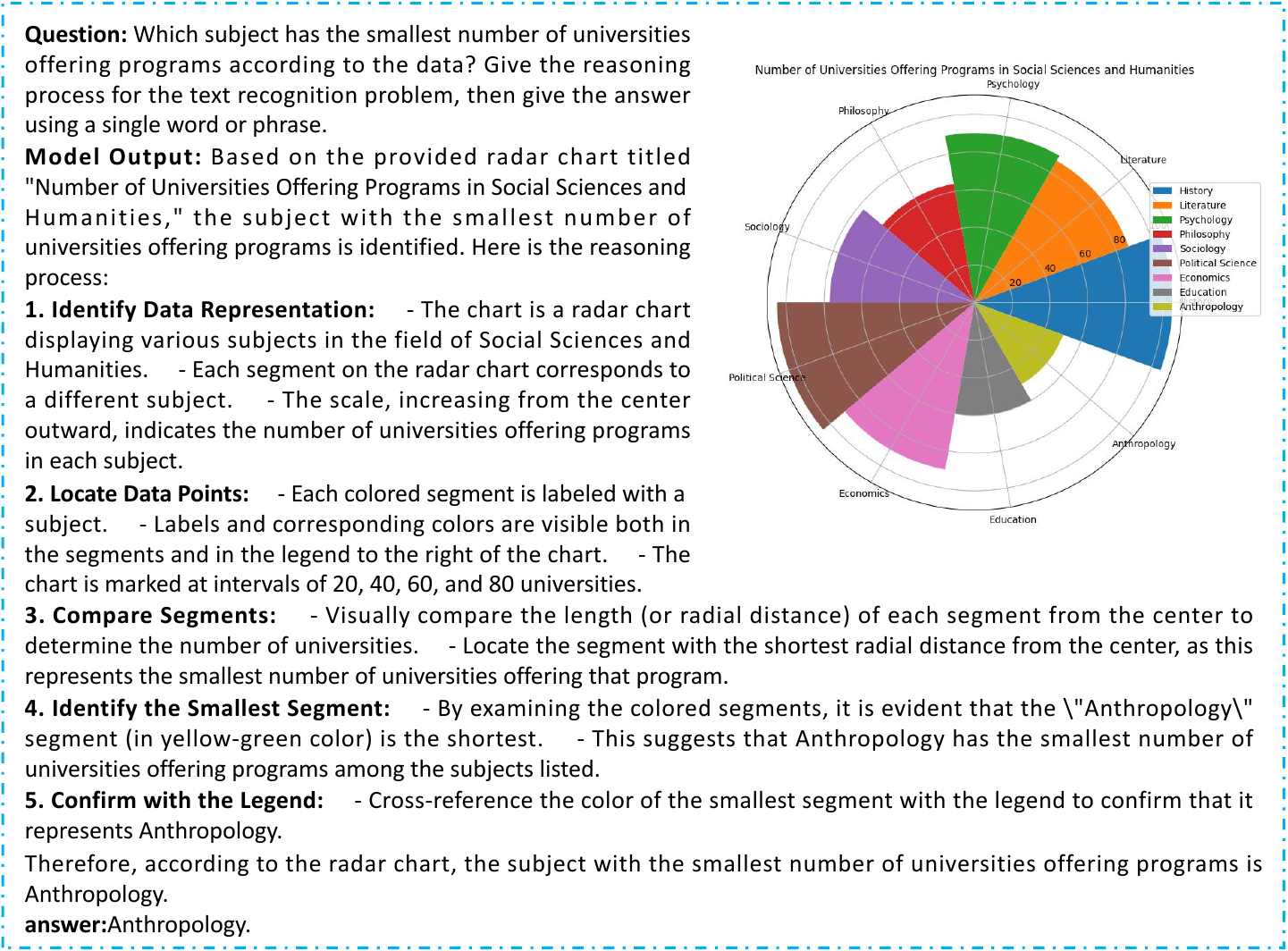}%
    }\hspace{2em}
     \subfloat[Science VQA]{%
        \includegraphics[width=0.4\linewidth]{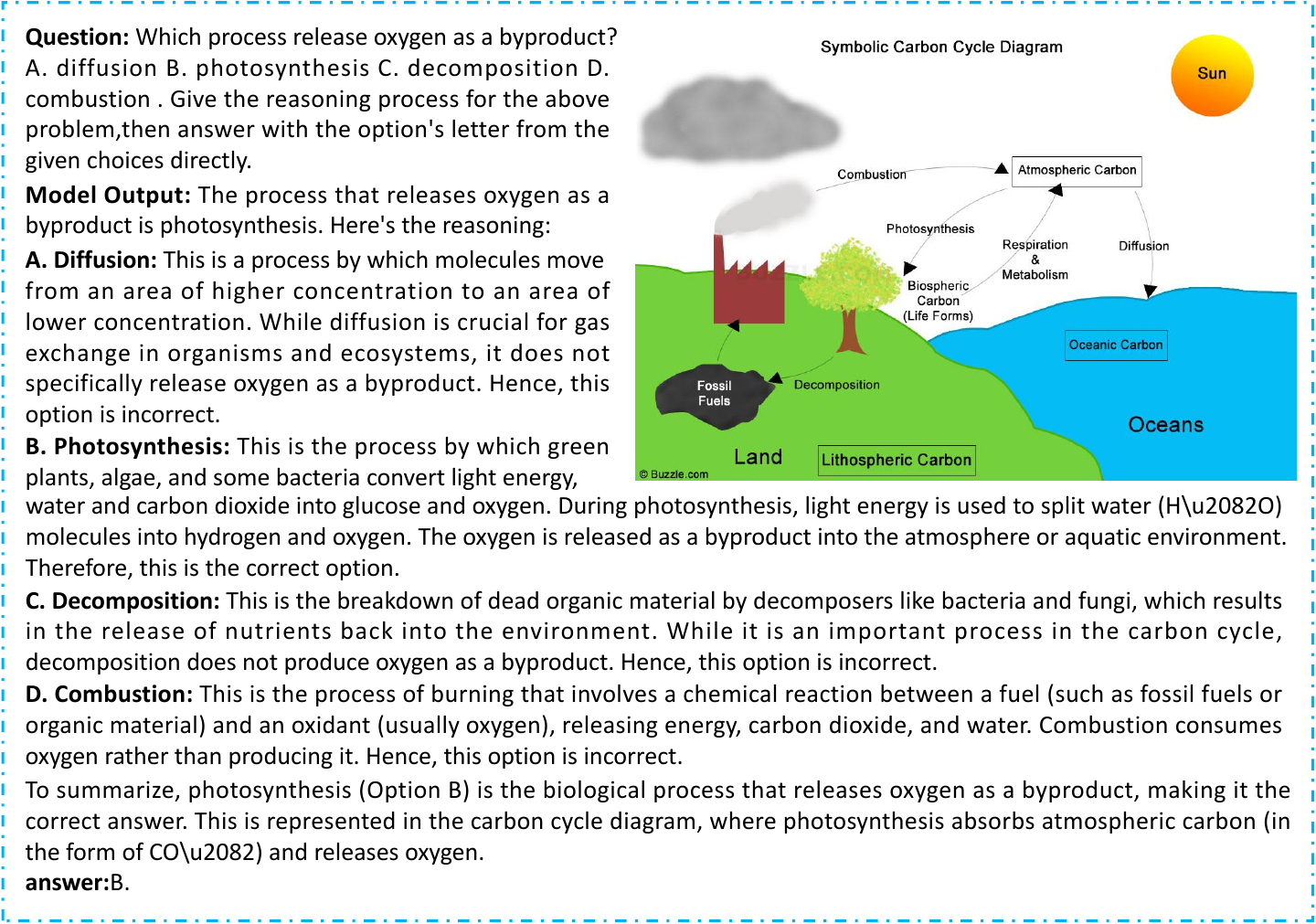}%
    }

    \vspace{0.7em}

    \subfloat[Medicine VQA]{%
        \includegraphics[width=0.4\linewidth]{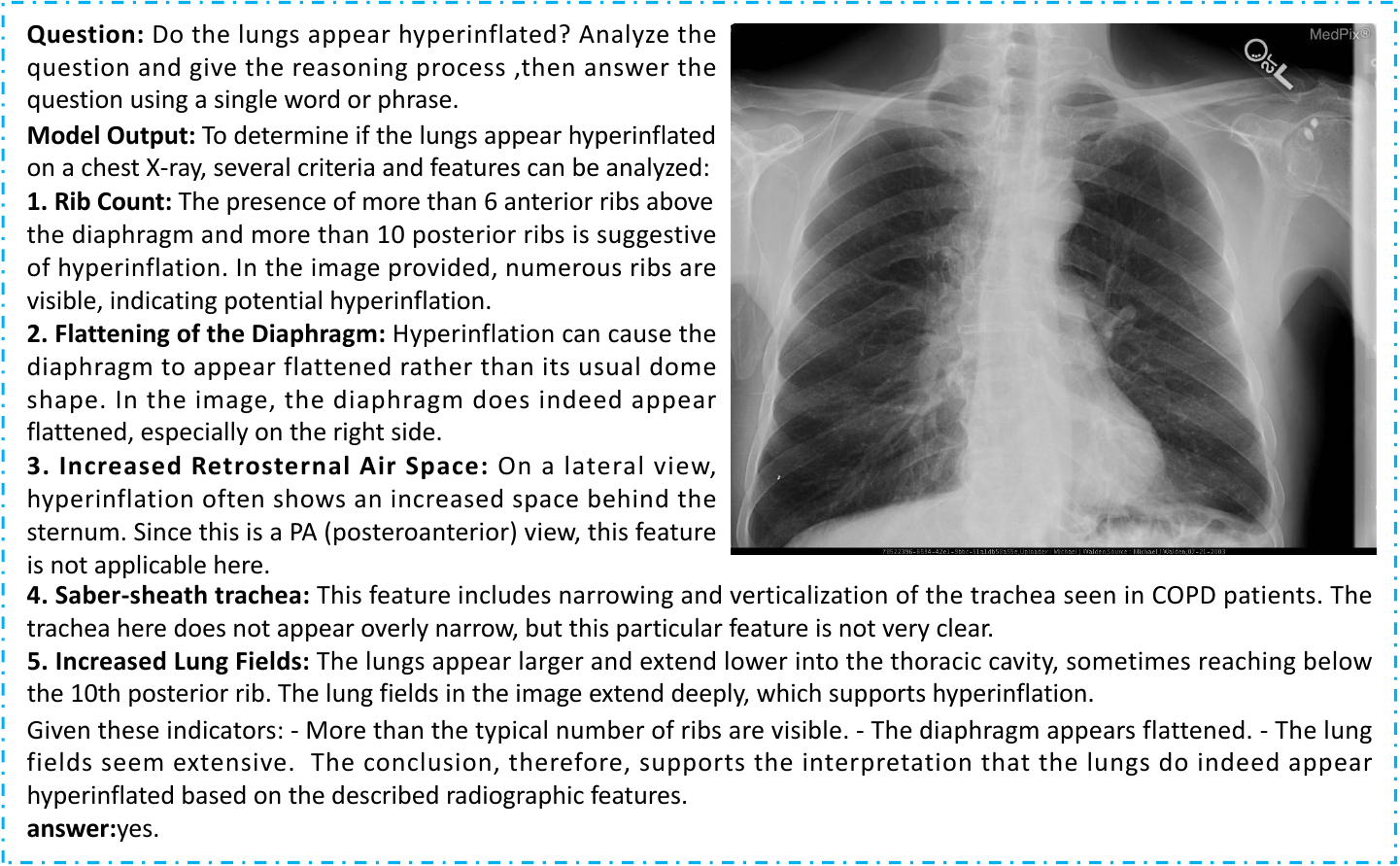}%
    }

    \caption{\textbf{Representative examples across 7 tasks in MLLM-CTBench.}
    Each panel shows the input image, the canonical instruction prompt, and the GPT-4 annotation for CoT reasoning process.}
    \label{fig:supp_examples}
\end{figure*}

\paragraph{Representative task examples.}
To make the benchmark composition and task diversity transparent, we provide one representative example per task in Figure~\ref{fig:supp_examples}, including the input image, the instruction prompt, and the GPT CoT reasoning process annotation. Each dataset poses distinct reasoning
challenges, ranging from mathematical derivation to visual
perception and domain-specific understanding.

\paragraph{Canonical instruction prompts and metrics.}
We unify the model interface by using one canonical instruction template per task while preserving the underlying task semantics.
Table~\ref{tab:supp_prompts_metrics} lists the instruction prompts and the metrics used for final-answer evaluation.
We adopt Exact Match for classification or QA
tasks requiring strict string alignment, and ROUGE-L for generation tasks to measure sequence overlap via longest common subsequence.
\begin{figure}[htbp]
    \centering
    \includegraphics[width=\linewidth]{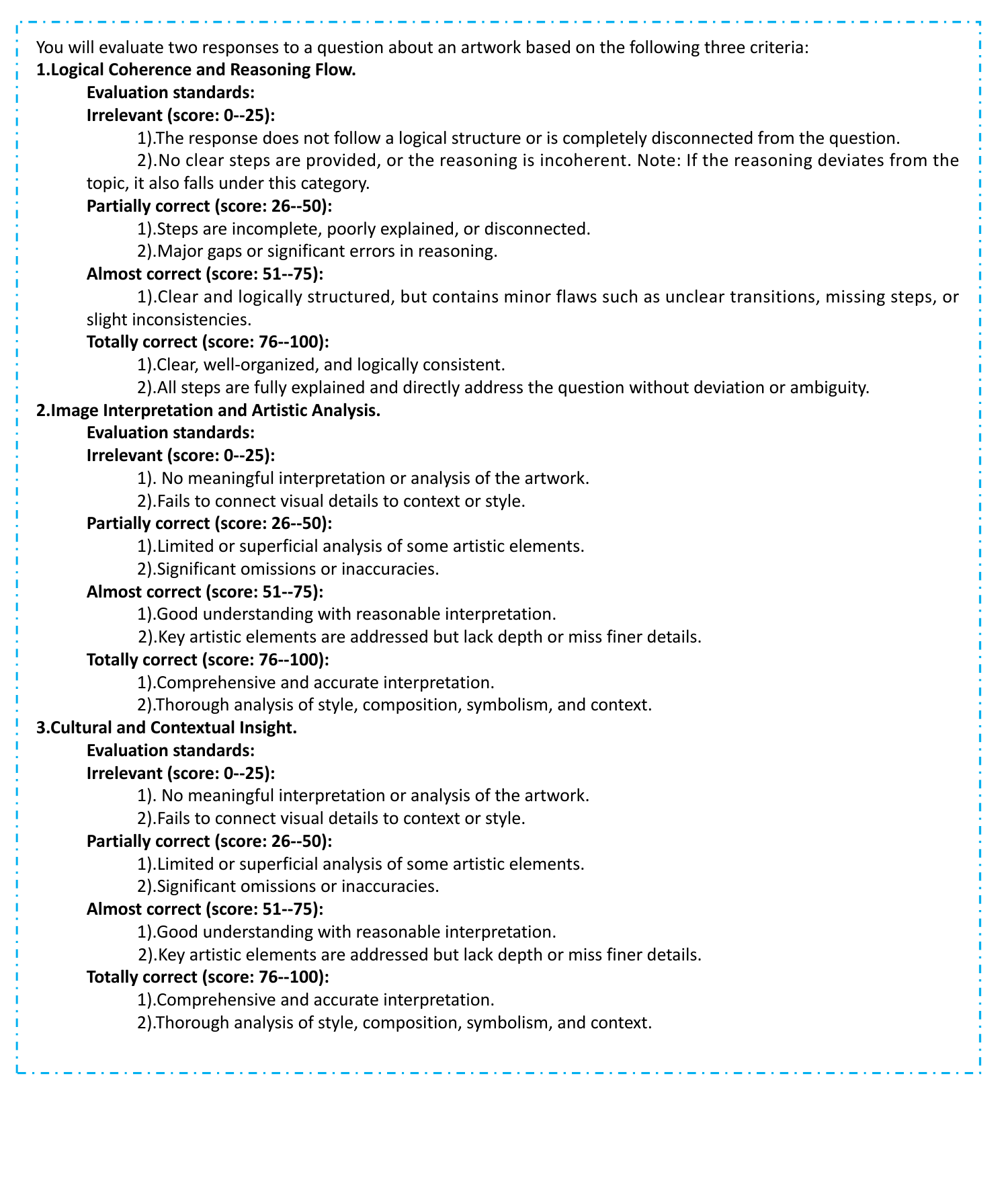}
    \caption{\textbf{Unified prompt} used to produce fine-grained CoT evaluation labels.}
    \label{fig:supp_rubric}
\end{figure}

\begin{table*}[t] 
\centering
\caption{Canonical instruction prompts and metrics across tasks.}
\label{tab:supp_prompts_metrics}
\setlength{\tabcolsep}{2.6mm}
\renewcommand{\arraystretch}{1.05}
\resizebox{\textwidth}{!}{%
\begin{tabular}{l l l}
\toprule
\textbf{Task} & \textbf{Instruction Prompt (Canonical)} & \textbf{Final-Answer Metric} \\
\midrule
Math QA & Solve the following math problem and give your reasoning, then give the answer. & Exact Match \\
Economics QA & Give your reasoning about the monetary policy stance, then answer with the option’s letter directly. & Exact Match \\
Science VQA & Give the reasoning process, then answer with the option’s letter directly. & Exact Match \\
Math VQA & Analyze the problem and give the solution; then answer with the option’s letter. & Exact Match / ROUGE-L \\
Medicine VQA & Analyze and give the reasoning process, then answer using a single word or phrase. & ROUGE-L \\
OCR VQA & Give the reasoning process for text recognition, then answer using a single word or phrase. & ROUGE-L \\
Arts VQA & Analyze the artwork and give a reasoning process, then answer briefly. & ROUGE-L \\
\bottomrule
\end{tabular}%
}
\end{table*} 


  \begin{table*}[t]
·
    \centering
    \caption{Evaluation of continual instruction tuning of MLLMs using macro-level metrics (final answer accuracy) on InternVL3-2B. Results are reported for three models under both Order-A and Order-B. For sequential finetuning, each cell reports Acc (top) and Forget (bottom).}
    \resizebox{\textwidth}{!}
    {

    \begin{tabular}{cc|*{7}{c}|cc}
      \hline
      \textbf{Model} & \textbf{Setting} 
      & \textbf{Math QA} & \textbf{Arts VQA} & \textbf{Math VQA} & \textbf{Econ. QA}
      & \textbf{Med. VQA} & \textbf{OCR VQA} & \textbf{Sci. VQA} & \textbf{AP} & \textbf{BWT} \\
      \hline

      \multirow{7}{*}{\textbf{InternVL3-2B}}
          &Multi-task  &88.92 & 27.13 & 66.17 & 36.89 & 35.06 & 42.16 & 87.94 & 54.90 & --\\
          &Zero-shot  &23.21 & 4.09 & 54.13 & 12.58 & 25.09 & 29.08 & 53.28 & 28.78  & --\\
          &Direct FT  &89.64 & 28.57 & 64.89 & 37.43 & 33.28 & 43.17 & 83.21 & 54.31 & --\\ \cline{2-11}
          & \makecell{Sequential Finetuning\\Order-A}
              & \makecell{46.55\\-42.37} & \makecell{9.49\\-18.10} & \makecell{35.49\\-29.96} & \makecell{34.59\\-1.02} & \makecell{27.57\\-5.21} & \makecell{39.12\\-5.47} & \makecell{84.54\\-} & 39.62 & \makecell{-14.59} \\ \cline{2-11}
         & \makecell{Sequential Finetuning\\Order-B}
            & \makecell{86.45\\-} & \makecell{7.14\\-19.81} & \makecell{37.79\\-25.04} & \makecell{37.22\\+1.00} & \makecell{27.21\\-4.15} & \makecell{24.36\\-19.66} & \makecell{81.24\\-5.00} & 43.06 & \makecell{-10.38} \\ \cline{2-11}
            
      \hline
    \end{tabular}
    } 

    \label{Internvl_baseline}
  \end{table*}

\section{Additional Backbone Results}
Due to space constraints, the main paper reports macro-level continual instruction tuning results on two representative backbones (LLaVA-1.5-7B and Qwen2.5-VL-3B). Here we further include InternVL3-2B as an additional backbone to verify the generality of our findings. We follow the same task suite, task orders (Order-A/Order-B), training protocol, and evaluation procedure as in the main paper. Consistent with our macro-level reporting, each entry in Table~\ref{Internvl_baseline} presents the final performance (Acc) and the corresponding Forget (difference between the post-task score and the final score. 
Overall, InternVL3-2B exhibits clear catastrophic forgetting and order-dependent interference, in line with the trends observed in the main results.
\section{Prompts and Reliability of the CoT Evaluator}
\label{sec:supp_evaluator}

\subsection{Unified rubric prompt for fine-grained CoT Annotation}
\label{sec:supp_rubric_prompt}

We adopt a unified rubric that scores Chain-of-Thought reasoning quality along complementary dimensions (e.g., logical coherence, visual grounding / interpretation, and domain knowledge).
Figure~\ref{fig:supp_rubric} shows the exact prompt template used to obtain fine-grained Annotations.

\subsection{A concrete evaluation example}
\label{sec:supp_eval_example}

\begin{figure*}[t]
    \centering
    \includegraphics[width=\linewidth]{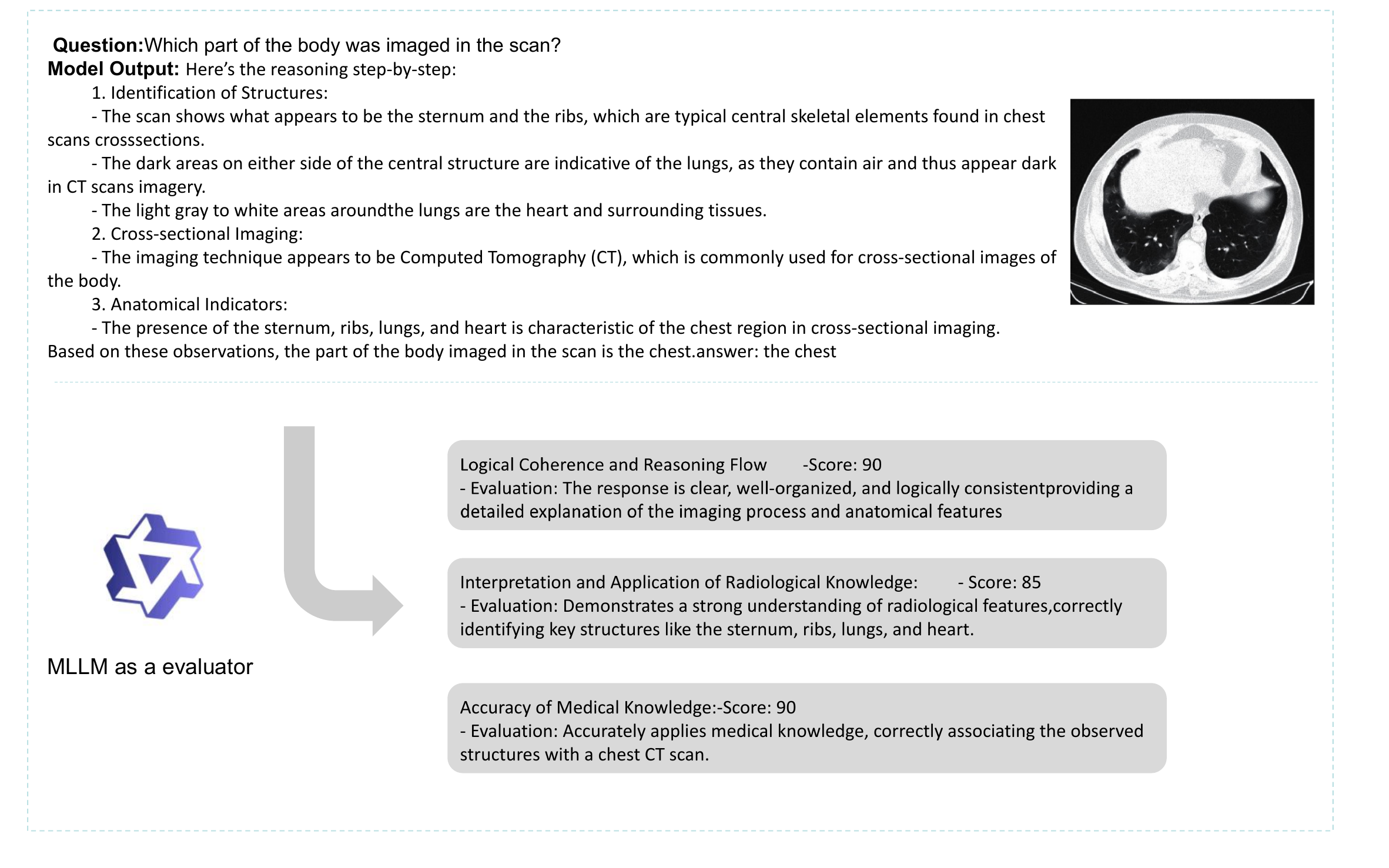}
    \caption{\textbf{Example of process-level evaluation} using MLLM-based evaluator.}
    \label{fig:supp_eval_example}
\end{figure*}

Figure~\ref{fig:supp_eval_example} illustrates one example of evaluator-based process assessment: given a model’s CoT, the evaluator assigns dimension-wise scores and aggregates them into a final CoT score.

\subsection{Correlation metrics for evaluator--human agreement}
\label{sec:supp_corr_metrics}

We quantify agreement between evaluator scores and human references using Spearman’s $\rho$, Pearson’s $r$, and Kendall’s $\tau$.

\paragraph{Pearson’s $r$.}
Pearson’s correlation coefficient measures linear association between two continuous variables:
\begin{equation}
r = \frac{\sum_{i=1}^{n} (x_i - \bar{x})(y_i - \bar{y})}
{\sqrt{\sum_{i=1}^{n} (x_i - \bar{x})^2}\;\sqrt{\sum_{i=1}^{n} (y_i - \bar{y})^2}},
\end{equation}
where $\bar{x}$ and $\bar{y}$ are sample means.

\paragraph{Kendall’s $\tau$.}
Kendall’s tau evaluates ordinal association between two rankings:
\begin{equation}
\tau = \frac{N_c - N_d}{\tfrac{1}{2} n (n - 1)},
\end{equation}
where $N_c$ and $N_d$ denote the numbers of concordant and discordant pairs.
Compared with Spearman’s $\rho$, Kendall’s $\tau$ is typically more conservative and less sensitive to large rank differences.

\section{Additional Continual Learning Results Under Order-B}
\label{sec:supp_orderB}

\begin{table*}[t]
\centering
\caption{Performance of representative continual learning methods with LLaVA-1.5-7B and Qwen2.5-VL-3B on MLLM-CTBench (Order-B), evaluated using macro-level final answer accuracy.}
\label{tab:method_both_orderB}

\subfloat[LLaVA-1.5-7B\label{tab:method_llava_orderB}]{
\begin{minipage}{\textwidth}
\centering
{\fontsize{9pt}{10pt}\selectfont
\setlength{\tabcolsep}{1mm}
\begin{tabular}{c*{7}{cc}cc}
\toprule
\multirow{2}{*}{\textbf{Method}} &
\multicolumn{2}{c}{\textbf{Math QA}} &
\multicolumn{2}{c}{\textbf{Arts VQA}} &
\multicolumn{2}{c}{\textbf{Math VQA}} &
\multicolumn{2}{c}{\textbf{Econ. QA}} &
\multicolumn{2}{c}{\textbf{Med. VQA}} &
\multicolumn{2}{c}{\textbf{OCR VQA}} &
\multicolumn{2}{c}{\textbf{Sci. VQA}} &
\multirow{2}{*}{\textbf{AP}} & \multirow{2}{*}{\textbf{BWT}} \\
\cmidrule(lr){2-3}\cmidrule(lr){4-5}\cmidrule(lr){6-7}\cmidrule(lr){8-9}\cmidrule(lr){10-11}\cmidrule(lr){12-13}\cmidrule(lr){14-15}
& Acc & Forget & Acc & Forget & Acc & Forget & Acc & Forget & Acc & Forget & Acc & Forget & Acc & Forget &  &  \\
\midrule
\textbf{ER} &
81.28 & -- &
27.51 & $+0.03$ &
42.42 & $-2.73$ &
65.32 & $-1.62$ &
28.38 & $-1.91$ &
17.28 & $-2.66$ &
71.91 & $-5.75$ &
47.73 & $-2.09$ \\
\textbf{DER} &
83.50 & -- &
30.56 & $+0.38$ &
46.07 & $+0.80$ &
70.26 & $+1.31$ &
30.10 & $-2.43$ &
21.44 & -- &
57.02 & $-2.83$ &
48.42 & $-0.40$ \\
\textbf{EWC} &
79.56 & -- &
13.67 & $-15.80$ &
22.01 & $-23.37$ &
61.09 & $-9.47$ &
14.78 & $-15.17$ &
13.32 & $-8.19$ &
50.42 & $-25.36$ &
36.41 & $-13.91$ \\
\textbf{MAS} &
68.72 & -- &
21.60 & $-4.03$ &
41.16 & $-2.74$ &
60.89 & $-6.65$ &
27.39 & $-2.12$ &
14.53 & $-4.42$ &
60.04 & $-17.72$ &
42.08 & $-5.35$ \\
\textbf{LwF} &
67.49 & -- &
12.22 & $-10.68$ &
27.14 & $-13.45$ &
58.87 & $-9.48$ &
23.81 & $-6.14$ &
10.97 & $-7.90$ &
46.56 & $-12.14$ &
35.29 & $-8.54$ \\
\textbf{freeze-init} &
81.28 & -- &
28.97 & $-0.16$ &
44.81 & $-0.80$ &
65.93 & $-4.03$ &
30.04 & $+3.76$ &
20.23 & $-1.21$ &
55.04 & $-2.83$ &
46.61 & $-0.75$ \\
\textbf{freeze-last} &
81.28 & -- &
28.64 & $-1.66$ &
41.51 & $-3.30$ &
69.66 & $-0.50$ &
29.94 & $+2.39$ &
19.59 & $-1.85$ &
57.68 & $-3.11$ &
46.90 & $-1.15$ \\
\textbf{L2P} &
64.17 & $-12.01$ &
27.68 & $-2.61$ &
40.96 & $-5.02$ &
57.61 & $-3.58$ &
22.95 & $-2.20$ &
14.58 & $-4.65$ &
53.96 & $-20.99$ &
40.27 & $-7.30$ \\
\textbf{MagMaX} &
41.38 & $-38.18$ &
12.35 & $-17.12$ &
34.78 & $-10.60$ &
66.13 & $-4.43$ &
23.13 & $-6.82$ &
17.31 & $-4.20$ &
62.30 & $-13.48$ &
36.77 & $-13.55$ \\
\bottomrule
\end{tabular}}
\end{minipage}
}

\par\vspace{6pt} 

\subfloat[Qwen2.5-VL-3B\label{tab:method_qwen_orderB}]{
\begin{minipage}{\textwidth}
\centering
{\fontsize{9pt}{10pt}\selectfont
\setlength{\tabcolsep}{1mm}
\begin{tabular}{c*{7}{cc}cc}
\toprule
\multirow{2}{*}{\textbf{Method}} &
\multicolumn{2}{c}{\textbf{Math QA}} &
\multicolumn{2}{c}{\textbf{Arts VQA}} &
\multicolumn{2}{c}{\textbf{Math VQA}} &
\multicolumn{2}{c}{\textbf{Econ. QA}} &
\multicolumn{2}{c}{\textbf{Med. VQA}} &
\multicolumn{2}{c}{\textbf{OCR VQA}} &
\multicolumn{2}{c}{\textbf{Sci. VQA}} &
\multirow{2}{*}{\textbf{AP}} & \multirow{2}{*}{\textbf{BWT}} \\
\cmidrule(lr){2-3}\cmidrule(lr){4-5}\cmidrule(lr){6-7}\cmidrule(lr){8-9}\cmidrule(lr){10-11}\cmidrule(lr){12-13}\cmidrule(lr){14-15}
& Acc & Forget & Acc & Forget & Acc & Forget & Acc & Forget & Acc & Forget & Acc & Forget & Acc & Forget &  &  \\
\midrule
\textbf{ER} &
94.09 & -- &
25.26 & $-6.23$ &
58.24 & $-11.89$ &
90.58 & $+2.63$ &
25.46 & $-7.53$ &
37.12 & $-9.51$ &
78.71 & $-9.98$ &
58.49 & $-6.08$ \\
\textbf{DER} &
94.83 & -- &
28.90 & $-5.97$ &
68.90 & $-2.86$ &
89.63 & $+3.61$ &
32.80 & $-1.84$ &
44.80 & $-5.32$ &
86.57 & $-3.55$ &
63.78 & $-2.27$ \\
\textbf{EWC} &
92.61 & -- &
20.39 & $-14.18$ &
67.39 & $-3.42$ &
70.36 & $+32.05$ &
28.88 & $-4.64$ &
32.46 & $-16.93$ &
80.02 & $-8.58$ &
56.02 & $-2.24$ \\
\textbf{MAS} &
96.55 & -- &
21.30 & $-12.69$ &
67.50 & $-4.56$ &
83.77 & $-4.03$ &
32.18 & $-1.36$ &
36.85 & $-12.56$ &
78.98 & $-8.96$ &
59.59 & $-6.31$ \\
\textbf{LwF} &
80.30 & -- &
29.65 & $+0.97$ &
67.16 & $+0.91$ &
77.92 & $-7.16$ &
29.35 & $-3.50$ &
38.93 & $-8.71$ &
80.77 & $-8.77$ &
57.73 & $-3.75$ \\
\textbf{freeze-init} &
89.68 & -- &
28.92 & $+0.15$ &
45.84 & $-15.62$ &
80.75 & $-9.01$ &
28.74 & $-3.94$ &
34.11 & $-9.74$ &
51.65 & $-20.26$ &
51.38 & $-8.35$ \\
\textbf{freeze-last} &
89.41 & -- &
25.74 & $-5.16$ &
65.68 & $-1.59$ &
76.59 & $-9.94$ &
27.52 & $-4.09$ &
30.33 & $-14.57$ &
75.31 & $-9.52$ &
55.80 & $-6.41$ \\
\textbf{L2P} &
79.48 & $-1.75$ &
30.13 & $-2.85$ &
65.48 & $-4.30$ &
76.98 & $-6.58$ &
28.95 & $-2.74$ &
39.17 & $-4.80$ &
79.88 & $-6.90$ &
57.40 & $-4.00$ \\
\textbf{MagMaX} &
95.07 & $+3.20$ &
10.53 & $-25.84$ &
70.24 & $-0.91$ &
92.54 & $+8.37$ &
32.33 & $-2.91$ &
42.59 & $-4.66$ &
83.79 & $-5.75$ &
61.01 & $-4.07$ \\

\bottomrule
\end{tabular}}
\end{minipage}
}

\end{table*}
We report the macro-level final-answer results under Order-B to complement the Order-A results in the main paper.


\section{Why KL-Regularized RFT Is More Stable: Distributional and Bayesian Views}
\label{sec:supp_rft_theory}

\paragraph{Distributional perspective.}
GRPO optimizes a KL-regularized expected reward objective:
\begin{equation}
\max_{\pi_\theta}\;\mathbb{E}_{\pi_\theta}[R] - \beta\,\mathrm{KL}(\pi_\theta \,\|\, \pi_{\text{ref}}),
\label{eq:supp_grpo}
\end{equation}
where $\pi_{\text{ref}}$ is a reference policy (the model checkpoint after continually learning on prior task) and $\beta$ controls regularization strength.
The KL term constrains distributional drift, promoting representational consistency and mitigating forgetting under sequential updates.

\paragraph{SFT contrast.}
SFT minimizes negative log-likelihood on supervised data:
\begin{equation}
\mathcal{L}_{\mathrm{SFT}} = -\mathbb{E}_{(x,y)\sim q}\log \pi_\theta(y|x).
\label{eq:supp_sft}
\end{equation}
When the supervision distribution $q(y|x)$ diverges from $\pi_{\text{ref}}(y|x)$, SFT can induce larger drift, increasing interference across tasks.

\paragraph{Bayesian analogy (interpretation).}
Beyond the geometric intuition, we provide a complementary probabilistic view based on approximate Bayesian inference. Let $\pi_{\text{ref}}(y|x)$ denote a prior belief and $q(y|x)$ the supervision-induced likelihood. Then the updates can be interpreted as:
\begin{align}
\pi_{\mathrm{SFT}}(y|x) &\propto q(y|x), \\
\pi_{\mathrm{RFT}}(y|x) &\propto \pi_{\text{ref}}(y|x)\exp\!\left(\frac{Q(y|x)}{\beta}\right),
\end{align}
where $Q(y|x)$ denotes the reward or preference score derived from CoT reasoning traces or final-answer feedback. This view is an analytical analogy rather than a formal derivation. From this perspective, SFT corresponds to a pure likelihood maximization process that may collapse the posterior around task-specific modes, undermining prior knowledge. In contrast, RFT performs a tempered posterior refinement, reinforcing relevant output modes while preserving global structure.

While the distributional geometry explains the stability of GRPO through KL regularization, the Bayesian view highlights how RFT modulates belief revision without fully overwriting the pretrained prior. These two interpretations offer complementary explanations of the same underlying mechanism, jointly clarifying the plasticity–stability trade-off under continual instruction tuning.

\paragraph{Joint training strategy.}
Given the complementary strengths of the two paradigms, we adopt a hybrid training strategy: SFT is first applied to facilitate rapid adaptation to new instruction distributions, followed by GRPO-based RFT to consolidate and regularize the updated policy within the pretrained model's intrinsic manifold. This protocol aligns with the continual instruction tuning scenario, where tasks are presented sequentially and the model must incrementally integrate new skills while preserving prior knowledge. Our empirical results (Table~\ref{GRPO_SFT}) confirm that this joint strategy offers a favorable balance between adaptability and retention across diverse instruction domains in MLLMs.

\vfill
\end{document}